\documentclass[journal]{IEEEtran}
\usepackage{changepage}
\usepackage{hyperref}
\usepackage{color}
\usepackage{comment}
\usepackage{cite}
\usepackage{amsmath,amsfonts,amssymb,amsthm,bm,mathrsfs,mathtools,stmaryrd,units}
\usepackage{enumitem}
\usepackage{cleveref}
\usepackage{float}
\usepackage{graphicx}
\usepackage[table,xcdraw]{xcolor}
\usepackage{readarray}
\usepackage{caption}
\usepackage{subcaption}
\usepackage{multirow}
\usepackage[ruled,vlined]{algorithm2e}
\usepackage{enumerate}
\usepackage{enumitem}

\usepackage{booktabs}

\usepackage{tikz}
\usetikzlibrary{arrows, positioning}
\usetikzlibrary{decorations.pathreplacing}
\usepackage{pgfplots}
\usetikzlibrary{calc}
\usepgfplotslibrary{groupplots,dateplot}
\pgfplotsset{compat=1.15} 
\newcommand\inputpgf[2]{{
\let\pgfimageWithoutPath\pgfimage
\renewcommand{\pgfimage}[2][]{\pgfimageWithoutPath[##1]{#1/##2}}
\input{#1/#2}
}}

\usepackage[ruled]{algorithm2e}

\usepackage{lipsum} 

\SetKwInput{KwParams}{Parameters}
\SetKwInput{KwHyperParams}{Hyperparameters}

\begin{document}

\title{Classification of Buried Objects from Ground Penetrating Radar Images by using Second Order Deep Learning Models}
\author{Douba Jafuno, Ammar Mian, Guillaume Ginolhac, {\it Senior Member, IEEE},  Nickolas Stelzenmuller
\thanks{Douba Jafuno, Ammar Mian and Guillaume Ginolhac are with LISTIC (EA3703), University Savoie Mont-Blanc, FRANCE. Nickolas Stelzenmuller is with Geolithe, FRANCE. This work has been done thanks to the facilities offered by the Univ. Savoie Mont-Blanc - CNRS/IN2P3 MUST computing center.}
}

\maketitle

\begin{abstract}
In this paper, a new classification model based on covariance matrices is built in order to classify buried objects. The inputs of the proposed models are the hyperbola thumbnails obtained with a classical Ground Penetrating Radar (GPR) system. \textcolor{blue}{These thumbnails are then inputs to the first layers of a classical CNN, which then produces a covariance matrix using the outputs of the convolutional filters.} Next, the covariance matrix is given to a network composed of specific layers to classify Symmetric Positive Definite (SPD) matrices. We show in a large database that our approach outperform shallow networks designed for GPR data and conventional CNNs typically used in computer vision applications, particularly when the number of training data decreases and in the presence of mislabeled data. We also illustrate the interest of our models when training data and test sets are obtained from different weather modes or considerations.
\end{abstract}


\section{Introduction}
	\label{sec:intro}
	The Ground Penetrated Radar (GPR) is a RADAR system that provides an image of the underground \cite{daniels04,jol09,Benedetto2017}. In particular, it can be used to image buried objects such as mines, pipes (metal, plastic, cast iron, etc.) and even cavities. The main drawback of GPR images is that they are very noisy, in particular due to the clutter that is the sum of all contributions from micro-scatterers in the ground but also because of the strong answer of the different layers of the ground. \textcolor{blue}{The buried objects are therefore often difficult to detect/locate and even more difficult to classify.} One solution is to use complex systems such as stepped frequency RADAR \cite{gurbuz09}, Multiple Input Multiple Output (MIMO) \cite{zeng15},  or polarimetric sensors. But the high cost of these devices is not always attractive for industrial or civil engineering applications. In this paper, we consider classical GPR systems emitting a single wave, called Ricker, and whose image is created by a displacement on one axis of the transmitting/receiving system. In the normal configuration, the RADAR is positioned very close to the ground. In our case, we will study the possibility of placing the RADAR at a certain height above the ground. This study will enable us to assess the robustness of our approach in the case of using GPR placed on a drone.

As noticed previously, the bad quality of the GPR image requires the use of various signal processing, image processing or machine learning techniques to achieve just the right detection and localization performance. In machine learning, it is possible to use deep learning techniques for denoising or inversion \cite{Liu2021,He2022,Dai2022,ni22}, auto-encoders for detection \cite{Bestagini2021} or pattern recognition approaches for localization of buried objects \cite{Maas2013}. In signal processing, it is possible to use statistical methods normally used in detection \cite{Brunzell1999,zoubir02,Ho2002,hoarau17}, particle filtering \cite{ng08}, Markov fields \cite{manandhar15} or algebraic algorithms \cite{kovalenko07,Li2015}. In image processing, most of methods are based on inversion \cite{Wang2018,Terrasse2016}, compressive sensing \cite{Ambrosanio2015} or dictionary learning \cite{Giovanneschi19}. A robust inversion method has been proposed in \cite{gallet2023}, which achieves good performance whatever the type of soil or buried object. All these works are essential for good object detection and localization, but will not suffice if we wish to classify them and thus determine their physical properties. In this paper, we are interested in this last step. \textcolor{blue}{Before introducing the proposed approach and those of the literature, we give the different needed assumptions: firstly all the buried objects are detected and correctly localized and secondly we have a certain amount of labeled data at our disposal, enabling us to develop supervised approaches.}

For the classification of buried objects from GPR images, a number of studies already exist, based either on classical signal processing techniques \cite{Shao13,Tivive17}, machine learning \cite{Shao10,zhou18} or deep networks \cite{besaw15,almaimani18,travassos20,elsaadouny20}. In all these algorithms, \textcolor{blue}{the classification} is based on the shape of the hyperbola (in both axes), which is partly related to the shape of the buried object and its electromagnetic properties. Unfortunately, the shape of the hyperbola also depends on elements completely independent of the object. In particular, the technical characteristics of the GPR (frequency, elevation) as well as the type of soil and the number of layers between the object and the ground have an enormous influence on this shape. The aim of this paper is to propose a high-performance \textcolor{blue}{and robust} approach that will remain effective in as many experimental configurations as possible. \textcolor{blue}{It seems complicated to achieve this robustness by developing a shallow CNN model. One solution is to use a very deep network, as is the case in computer vision. But unfortunately, these networks require a large amount of training data to achieve good performance in terms of robustness. And it is well known that this large amount of data is not readily available for applications such as GPR.}

One solution is then to change of features before the classification step. Instead of the image of the hyperbola, a suitable transformation can achieve the performance and robustness objectives sought in this study. For example, it is possible to construct a covariance matrix from this image by using the method proposed in \cite{akodad2020ensemble}. It is known that second order can improve classification performance, as for example in computer vision \cite{li2017second,li2018towards}. Moreover, we have shown in \cite{Collas2023} that this kind of feature brings robustness when a shift is present between the training and the test data. By using a similar approach of \cite{akodad2020ensemble}, we have shown in a previous work \cite{gallet2022classification} that this operation achieves better performance with classical machine learning algorithms than using the raw image. This covariance matrix is constructed from the outputs of the first layers of a deep network. It measures the correlations between these different layers for a given image. To achieve a certain richness in this covariance matrix, it is often useful to use a certain number of layers (around 8-10). In this case, however, the covariance matrix is very large, making classification impossible caused by singularities issues. To solve this issue, a preliminary work \cite{huang2017riemannian} has proposed specific layers for Symmetric Positive Definite (SPD) matrices, properties of covariance matrices, to solve this problem of classifying from large covariance matrices. Similar models have also been proposed for EEG data analysis \cite{Kobler2022}, classification of RADAR data \cite{brooks2019riemannian} or classification in polarimetric SAR images \cite{shi2023riemannian}.

In this paper, we propose a new classification model based on a transformation of the raw image into a covariance matrix and specific subsequent layers adapted to this SPD matrix. We also propose a different approach to that proposed in \cite{akodad2020ensemble} for constructing our covariance matrix, which saves memory space while preserving correlation information. To train and to test our new model, we have from Geolithe, a sufficient database containing 4 types of buried objects with different GPR configurations (frequency and elevation) as well as several terrains (dry and wet sand and gravel). We compare our approach with shallow networks and conventional deep networks used in computer vision. We will show good performance and, above all, robustness of our pipeline to different experiments.

The outline of the paper is the following. First, section II introduces the GPR principle as well as some physical considerations allowing to better understand how the shape of the hyperbolas and the buried object are linked. Section III presents the new model to classify buried objects from GPR images. Next, section IV gives some details on the used database for the training and the steps. Finally, our approach is tested and compared to other algorithms in the section V. 

\section{Ground Penetrating Radar (GPR)}
	\label{sec:gpr}
	\subsection{GPR Principle}

\begin{figure}[t]
		\centering
\begin{tikzpicture}

\definecolor{darkgray176}{RGB}{176,176,176}
\definecolor{steelblue31119180}{RGB}{31,119,180}

\begin{axis}[
width=0.95\columnwidth,
height=4cm,
tick align=outside,
tick pos=left,
x grid style={darkgray176},
xlabel={Time (ns)},
xmin=-4.975, xmax=104.475,
xtick style={color=black},
y grid style={darkgray176},
ylabel={Amplitude},
ymin=-0.518573017580566, ymax=1.07231300083717,
ytick style={color=black}
]
\addplot [semithick, steelblue31119180]
table {%
0 -0.000969251586187207
0.5 -0.00115487174699431
1 -0.00137299856182274
1.5 -0.00162870723988541
2 -0.00192774696400008
2.5 -0.00227660887733857
3 -0.00268259713018404
3.5 -0.00315390239535746
4 -0.00369967708960488
4.5 -0.00433011134788404
5 -0.00505650858876988
5.5 -0.00589135928333985
6 -0.00684841129881528
6.5 -0.00794273493461814
7 -0.00919078050590108
7.5 -0.0106104260624973
8 -0.0122210125650677
8.5 -0.014043363581464
9 -0.0160997863224722
9.5 -0.0184140506156524
10 -0.021011342228416
10.5 -0.0239181868071028
11 -0.027162340608723
11.5 -0.03077264417784
12 -0.034778835174768
12.5 -0.0392113167048952
13 -0.0441008777442936
13.5 -0.0494783626150238
14 -0.0553742869448008
14.5 -0.0618183981586137
15 -0.0688391793011612
15.5 -0.0764632958828021
16 -0.0847149864793907
16.5 -0.0936153989956631
17 -0.103181875816664
17.5 -0.113427192511583
18 -0.124358756304152
18.5 -0.135977772163335
19 -0.148278386072111
19.5 -0.161246816770417
20 -0.174860489005109
20.5 -0.18908718301516
21 -0.203884216589378
21.5 -0.219197677508739
22 -0.234961725474708
22.5 -0.251097983675898
23 -0.26751504090451
23.5 -0.284108085548512
24 -0.300758692804924
24.5 -0.317334786037384
25 -0.333690792296469
25.5 -0.349668010600412
26 -0.365095209612574
26.5 -0.379789468836646
27 -0.393557274379591
27.5 -0.406195876718346
28 -0.417494913776033
28.5 -0.42723829800965
29 -0.43520636119213
29.5 -0.441178245211697
30 -0.44493452160017
30.5 -0.446260016743396
31 -0.444946813938315
31.5 -0.440797397770309
32 -0.43362790082785
32.5 -0.423271407691236
33 -0.409581266572584
33.5 -0.392434355088334
34 -0.371734243550779
34.5 -0.3474141970001
35 -0.319439956077762
35.5 -0.287812236863493
36 -0.252568891037691
36.5 -0.213786670240536
37 -0.171582542302422
37.5 -0.126114512111569
38 -0.0775819062261705
38.5 -0.0262250878601447
39 0.0276754225292944
39.5 0.0838004362816162
40 0.141794200108251
40.5 0.201267108201046
41 0.261799005587867
41.5 0.322943054024335
42 0.384230120391098
42.5 0.445173636605836
43 0.505274869713882
43.5 0.56402853134085
44 0.620928647313165
44.5 0.675474601197387
45 0.727177259971307
45.5 0.775565086182901
46 0.820190138905581
46.5 0.860633865647859
47 0.896512589167425
47.5 0.927482596873285
48 0.953244746128175
48.5 0.973548506194368
49 0.988195366662382
49.5 0.997041552785684
50 1
50.5 0.997041552785684
51 0.988195366662382
51.5 0.973548506194368
52 0.953244746128175
52.5 0.927482596873286
53 0.896512589167425
53.5 0.860633865647859
54 0.820190138905581
54.5 0.775565086182901
55 0.727177259971308
55.5 0.675474601197388
56 0.620928647313165
56.5 0.56402853134085
57 0.505274869713882
57.5 0.445173636605836
58 0.384230120391098
58.5 0.322943054024335
59 0.261799005587867
59.5 0.201267108201046
60 0.141794200108252
60.5 0.083800436281617
61 0.0276754225292952
61.5 -0.0262250878601439
62 -0.0775819062261698
62.5 -0.126114512111568
63 -0.171582542302422
63.5 -0.213786670240536
64 -0.252568891037691
64.5 -0.287812236863493
65 -0.319439956077762
65.5 -0.3474141970001
66 -0.371734243550779
66.5 -0.392434355088334
67 -0.409581266572584
67.5 -0.423271407691236
68 -0.43362790082785
68.5 -0.440797397770309
69 -0.444946813938315
69.5 -0.446260016743396
70 -0.44493452160017
70.5 -0.441178245211697
71 -0.43520636119213
71.5 -0.42723829800965
72 -0.417494913776033
72.5 -0.406195876718346
73 -0.393557274379592
73.5 -0.379789468836646
74 -0.365095209612575
74.5 -0.349668010600412
75 -0.33369079229647
75.5 -0.317334786037384
76 -0.300758692804924
76.5 -0.284108085548512
77 -0.26751504090451
77.5 -0.251097983675898
78 -0.234961725474708
78.5 -0.219197677508739
79 -0.203884216589378
79.5 -0.18908718301516
80 -0.174860489005109
80.5 -0.161246816770417
81 -0.148278386072112
81.5 -0.135977772163335
82 -0.124358756304152
82.5 -0.113427192511583
83 -0.103181875816664
83.5 -0.0936153989956631
84 -0.0847149864793907
84.5 -0.0764632958828021
85 -0.0688391793011612
85.5 -0.0618183981586137
86 -0.0553742869448008
86.5 -0.0494783626150238
87 -0.0441008777442936
87.5 -0.0392113167048952
88 -0.0347788351747682
88.5 -0.0307726441778401
89 -0.0271623406087231
89.5 -0.0239181868071028
90 -0.0210113422284161
90.5 -0.0184140506156524
91 -0.0160997863224722
91.5 -0.014043363581464
92 -0.0122210125650677
92.5 -0.0106104260624973
93 -0.0091907805059011
93.5 -0.00794273493461815
94 -0.00684841129881529
94.5 -0.00589135928333986
95 -0.00505650858876989
95.5 -0.00433011134788404
96 -0.00369967708960488
96.5 -0.00315390239535746
97 -0.00268259713018404
97.5 -0.00227660887733857
98 -0.00192774696400008
98.5 -0.00162870723988541
99 -0.00137299856182274
99.5 -0.00115487174699431
};
\end{axis}

\end{tikzpicture}
\caption{\footnotesize Waveform emitted by classical GPR, called Ricker.}
		\label{fig:ricker}	
\end{figure}
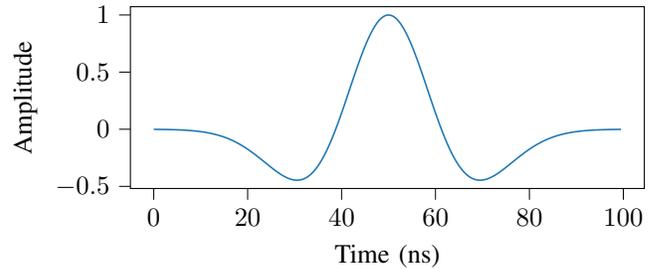

\begin{figure}[t]
\begin{center}
\includegraphics[width=3.53in, height=1.6in]{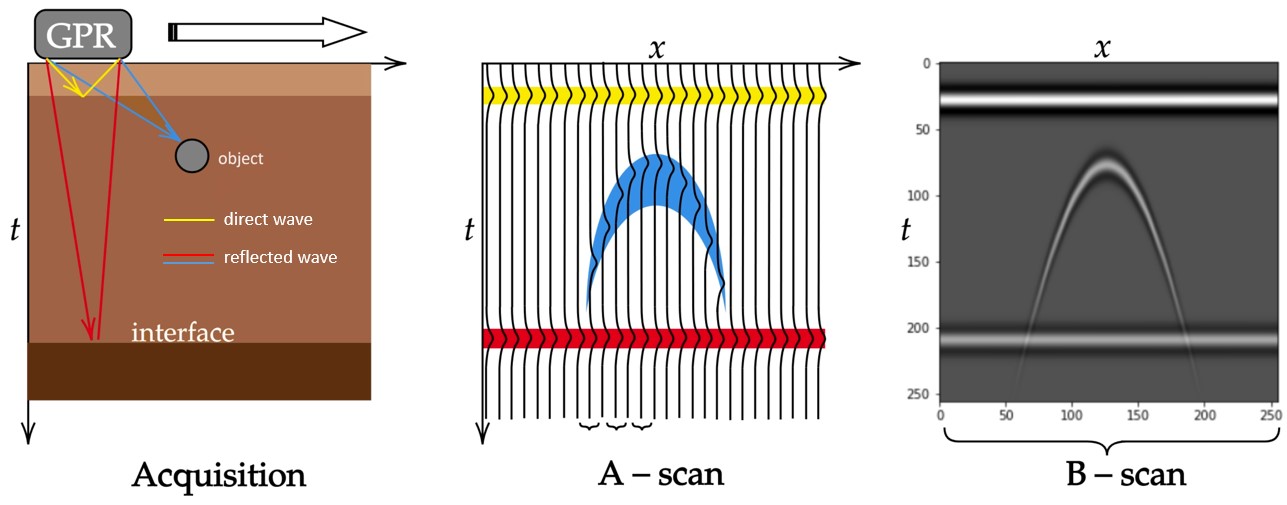}
\end{center}
\caption{GPR Principle with acquisition (left), creation of the A-scan (middle) and the B-scan (right).}
\label{fig:radargramme}
\end{figure}

Ground Penetrating Radar (GPR) is a radar system consisting of an antenna that is typically placed on the surface of the ground\footnote{\textcolor{blue}{In our paper, we will also consider different elevations but the principle presented applies still.}}. For a wide range of systems, the electromagnetic wave transmitted by the GPR is a simple wavelet, known in the community as a Ricker. An example of this type of signal is shown in figure \ref{fig:ricker}. The frequency of the wave, from 10MHz to 2GHz, depends on the application. In fact, this frequency is linked to the depth that can be reached. For example, for mine detection, a high frequency will be chosen, as there is no need to reach great depths, whereas the opposite will be chosen if you want to know the composition of the ground over several tens of meters.

The GPR is moved along an axis, as shown on the left of the figure \ref{fig:radargramme}. All acquisitions, i.e. the amplitude of the signal over time at a given point, are combined to form an image known as the A-scan.  From this A-scan, it is then possible to construct an image of the ground called a B-scan or radargram. All processing, from detection to classification, \textcolor{blue}{are} classically based on this image. If a buried object is present, it is then seen several times by the GPR, leading to a hyperbola in the B-scan image. Soils are often composed of several layers of different types. In this case, the B-scan shows some lines to represent these layers. On the right-hand side of the figure \ref{fig:radargramme}, the B-scan shows the image of a buried object and two layers. In this simple simulation, we do not take noise into account. In reality, the signal-to-noise ratio of GPR images is very low. In particular, there is a lot of clutter due to the Ricker reflecting off small scatterers, such as rocks.

In the next section, we will look at how object type and other parameters influence the shape of the hyperbola.

\subsection{Influence of the physical parameters on the hyperbola shape}

Several factors influence the shape of the hyperbola in the radargram. Firstly, GPR parameters, such as the frequency of the transmitted wave or the elevation of the system relative to the ground, strongly affect the resulting hyperbola. In this article, we will consider several frequencies as well as different GPR elevations. This last point is useful in the case of an airborne GPR, which cannot then be used too close to the ground. 

Another factor influencing the shape is obviously the soil and its composition. Each type of soil has its own dielectric permittivity and electrical conductivity, both of which influence the speed of the emitted wave. For example, higher dielectric permittivity generally slows down the propagation of electromagnetic waves or deflects radar waves more than soils with lower permittivity. \textcolor{blue}{Furthermore, interfaces between different layers also have an important influence, in part when the two layer parameters are very different from each other. In this case, the Ricker wavelet will undergo several deformations (attenuation, spectral broadening, etc.). For more details on these deformations, please refer to \cite{daniels04}.}

Finally, the shape of the hyperbola obviously depends on the buried object reflecting the transmitted wave back to the RADAR. Firstly, the size and shape of the object will influence its shape, particularly in the axis of motion of the RADAR. Finally, the electromagnetic properties of the buried object have an impact on the hyperbola, but more in the time axis.  Conductive materials, such as metals, absorb more energy and attenuate signals faster than non-conductive materials. To help distinguish between metallic and non-metallic objects, we can also look at the polarization of the hyperbola. Polarity reversal occurs when radar waves reflect off objects whose permittivity is higher than that of the surrounding medium (soil has a lower permittivity than metallic objects). To define polarity, we define that the black areas of the hyperbola correspond to a negative polarity '-' and the lighter or white areas to a positive polarity '+'. So, depending on the polarity of the incident wave, a metal object can appear as a positive (+ - +) or negative (- + -) reflection. 

We will show some examples of buried object images in the next section.

\subsection{Examples}

\begin{figure}[h]
    \begin{subfigure}{0.3\columnwidth}
        \centering
        \includegraphics[width=\textwidth, height=3cm]{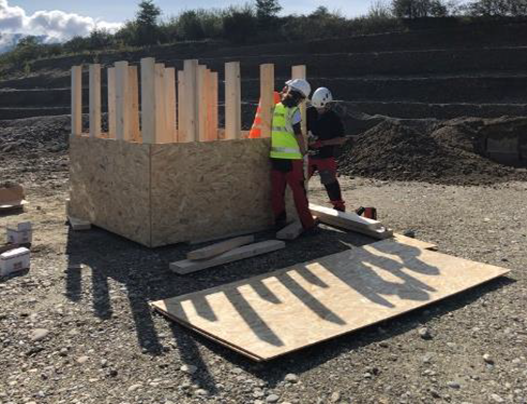}
        \label{fig:first-example}
    \end{subfigure}
    \begin{subfigure}{0.3\columnwidth}
        \centering
        \includegraphics[width=\textwidth, height=3cm]{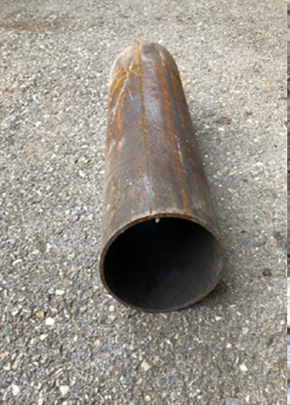}
        \label{fig:second-example}
    \end{subfigure}
    \begin{subfigure}{0.3\columnwidth}
        \centering
        \includegraphics[width=\textwidth, height=3cm]{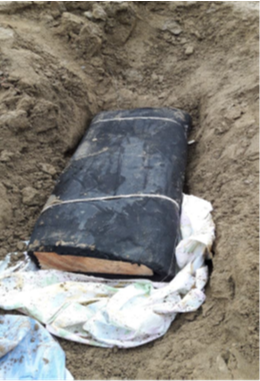}
        \label{fig:third-example}
    \end{subfigure}
    \\
    \begin{subfigure}{0.3\columnwidth}
        \centering
        \includegraphics[width=\columnwidth, height=3cm]{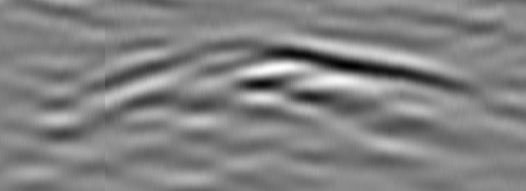}
        \caption{}
    \end{subfigure}
    \begin{subfigure}{0.3\columnwidth}
        \centering
        \includegraphics[width=\textwidth, height=3cm]{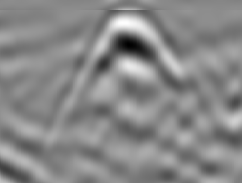}
        \caption{}
    \end{subfigure}
    \begin{subfigure}{0.3\columnwidth}
        \centering
        \includegraphics[width=\textwidth, height=3cm]{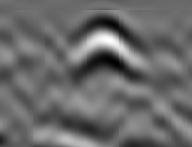}
        \caption{}
    \end{subfigure}
    \begin{subfigure}{0.05\columnwidth}
        \centering
        \includegraphics[width=\textwidth, height=3.cm]{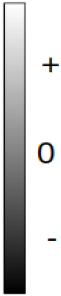}
        \caption{}
    \end{subfigure}
    \caption{Examples of preprocessed GPR images with the 200 MHz antenna in wet sand (after direct wave suppression and histogram correction thanks to GPRpy which is an open-source Ground Penetrating Radar processing and visualization software available in \url{https://github.com/NSGeophysics/GPRPy}) for different buried objects: Wooden shelter, dummy shell (Metal) and wooden board coated with rubber (Non-Metal). Noticed that the 3 radargrams have different scales.}
    \label{fig:examples}
\end{figure}

On the radargrams in Figure \ref{fig:examples}, we have used the 200 MHz GSSI antenna which have a positive polarity (+ - +) on wet sand. So a change in polarity will be detected as negative or reversed (- + -). It is therefore more appropriate to speak in terms of normal polarity when the reflection polarity is the same as the incident wave, or reversed polarity when these polarities are different.   

On the radargram (a), visible in Figure \ref{fig:examples}, the signature of the shelter consists of:
\begin{itemize}
    \item A wide hyperbola corresponding to the roof of the shelter (flat surface). The width of the flattened, high-intensity part corresponds to the width of the shelter (2 m). This reflection specifically corresponds to the soil-air interface (rather than soil-wood) due to the significant contrast in dielectric permittivity between air and soil and the thinness of the wood. The polarity is normal ((+ - +) here) same like GSI antenna : there is no change in polarity at the soil-air contact (decrease in dielectric permittivity).
    \item Two adjacent hyperbolas beneath the first reflection. They correspond to reflections on the corners of the shelter. Their polarity is inverse.
    \item A wide hyperbola below the previous hyperbolas, corresponding to the base of the shelter.
\end{itemize}

On the radargram (b), \textcolor{blue}{we observe that the signature of a dummy shell gives}
\begin{itemize}
    \item \textcolor{blue}{hyperbolas which are easily recognized by their high intensity, well-defined hyperbolic shape and polarity opposite that of GSSI antennas (- + -). The polarity is reversed for metallic objects.}
\end{itemize}

On the radargram (c), \textcolor{blue}{we observe that the signature of a wooden board coated with rubber gives} \textcolor{blue}{a slightly flattened hyperboles identified by a low intensity. Moreover, the polarity are reversed (- + -) here.}

\subsection{Goal of the classification model}

The 3 previous examples show that the shape of the hyperbola is linked to the buried object. In this article, we therefore propose to build a classification model that takes as input a thumbnail of each hyperbola. We assume that localization and classification have been carried out in the previous steps. Some pre-processing steps are given in \cite{gallet2022classification}.  

One of the main concerns regarding the classification strategy based on GPR images is robustness. Indeed, we noted in the previous sections that the shape of the hyperbola also depends on parameters other than those of the buried object, such as the GPR and the soil. In this case, the classification must be invariant to these deformations of the hyperbola independently of the object.

The proposed strategy in the next section consists in imposing a transformation on the thumbnail image in order to classify it after this treatment. In particular, we study the interest of the second-order model for achieving good classification performance with good robustness, and all this with relatively little training data.
	
\section{Second Order Model}
	\label{sec:som}
	
In our task of classifying our radargrams, we rely on neural networks that have shown their effectiveness in image classification tasks.\\




Let us first describe some of the models that we used as a baseline and illustrate their respective limitations. From this we will then describe our proposed approach that \textcolor{blue}{combines} several aspects of those models.

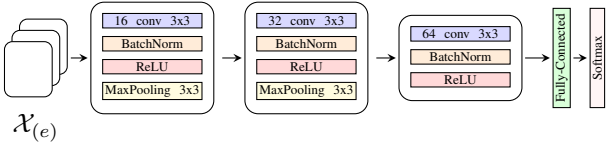
\begin{figure}[t]
  \centering
  \begin{tikzpicture}

  \node[draw, fill=white, rounded corners=3, text width=0.4cm, text height=0.5cm] (input1) {};
  \node[draw, fill=white, rounded corners=3, text width=0.4cm, text height=0.5cm,
        below left=0.15cm and 0.1 cm of input1, anchor=south west] (input2) {};
  \node[draw, fill=white, rounded corners=3, text width=0.4cm, text height=0.5cm,
        below left=0.15cm and 0.1 cm of input2, anchor=south west] (input3) {};
        \node[below= 0.25cm of input2] (batch_text) {$\mathcal{X}_{(e)}$};

  \begin{scope}
  \node[draw, rounded corners=5, text width=1.4cm, text height=1.25cm,
        right=0.4cm of input2] (layer1container){};
  \draw[-stealth, shorten >=1pt, shorten <= 4pt] (input2) -- (layer1container);
  \node[draw, below = 0.15cm of layer1container.north, text width=1.25cm,
        align=center, inner sep=1pt, fill=blue!15] (l1conv){\tiny 16 conv 3x3};
  \node[draw, below = 0.1cm of l1conv.south, text width=1.25cm, anchor=north,
        align=center, inner sep=1pt, fill=orange!15] (l1bn) {\tiny BatchNorm};
  \node[draw, below = 0.1cm of l1bn.south, text width=1.25cm, anchor=north,
        align=center, inner sep=1pt, fill=red!15] (l1relu) {\tiny ReLU};
  \node[draw, below = 0.1cm of l1relu.south, text width=1.25cm, anchor=north,
        align=center, inner sep=1pt, fill=yellow!15] (l1maxpool) {\tiny MaxPooling 3x3};
  \end{scope}

  \begin{scope}
  \node[draw, rounded corners=5, text width=1.4cm, text height=1.25cm,
        right=0.4cm of layer1container] (layer2container){};
  \draw[-stealth, shorten >=1pt, shorten <= 3pt] (layer1container) -- (layer2container);
  \node[draw, below = 0.15cm of layer2container.north, text width=1.25cm,
        align=center, inner sep=1pt, fill=blue!15] (l2conv){\tiny 32 conv 3x3};
  \node[draw, below = 0.1cm of l2conv.south, text width=1.25cm, anchor=north,
        align=center, inner sep=1pt, fill=orange!15] (l2bn) {\tiny BatchNorm};
  \node[draw, below = 0.1cm of l2bn.south, text width=1.25cm, anchor=north,
        align=center, inner sep=1pt, fill=red!15] (l2relu) {\tiny ReLU};
  \node[draw, below = 0.1cm of l2relu.south, text width=1.25cm, anchor=north,
        align=center, inner sep=1pt, fill=yellow!15] (l2maxpool) {\tiny MaxPooling 3x3};
  \end{scope}

  \begin{scope}
  \node[draw, rounded corners=5, text width=1.4cm, text height=0.9cm,
        right=0.4cm of layer2container] (layer3container){};
  \draw[-stealth, shorten >=1pt, shorten <= 3pt] (layer2container) -- (layer3container);
  \node[draw, below = 0.15cm of layer3container.north, text width=1.25cm,
        align=center, inner sep=1pt, fill=blue!15] (l3conv){\tiny 64 conv 3x3};
  \node[draw, below = 0.1cm of l3conv.south, text width=1.25cm, anchor=north,
        align=center, inner sep=1pt, fill=orange!15] (l3bn) {\tiny BatchNorm};
  \node[draw, below = 0.1cm of l3bn.south, text width=1.25cm, anchor=north,
        align=center, inner sep=1pt, fill=red!15] (l3relu) {\tiny ReLU};
  \end{scope}

  \node[draw, right=0.4cm of layer3container,
        text width=1.25cm, align=center, inner sep=1pt, anchor=north,
        fill=green!15, rotate=90] (FC){\tiny Fully-Connected};
  \draw[-stealth, shorten >=1pt, shorten <= 1pt] (layer3container.east) -- (FC.north);
  \node[draw, right=0.25cm of FC.south, anchor=north,
        text width=1.25cm, align=center, inner sep=1pt, 
        fill=pink!15, rotate=90] (softmax){\tiny Softmax};
  \draw[-stealth, shorten >=1pt, shorten <= 1pt] (FC.south) -- (softmax);

\end{tikzpicture}
  \caption{Architecture of model CNN1 \cite{almaimani18}}
  \label{fig: CNN1 architecture}
\end{figure}


\subsection{Shallow CNN network}

As we are considering GPR images thumbnails of pre-localized hyperbolas, the first classes of models that are natural to consider are 2D Convolutional Neural Networks (CNNs) that have been successfully applied in computer vision tasks \cite{krizhevsky2012imagenet}. Rather than consider large models, and given that we consider very few classes compared to the general image classification problem, shallow CNNs have been considered in \cite{almaimani18}, where several of such small scale architectures are presented. From this study, we selectioned the best reported model in terms of overall accuracy that is denoted \textcolor{blue}{\textbf{S-CNN} (Shallow CNN)} whose architecture is reported in Figure \ref{fig: CNN1 architecture}.

It is constructed as a succession of 3 embedding layers with help of standard 2D convolutions, batch normalization, ReLU non-linearity and max-pooling layers. Then a fully connected layer is followed by a softmax for classes probabilities. While this model allows for classification of the hyperbolas, the obtained accuracies we obtained in practice were not as satisfactory on our dataset\footnote{as will be seen in section \ref{sec:num_exp}.}, than in the one used in \cite{almaimani18}, who was trained on synthetic dataset and with a different number of classes. Since our database consists of a great number of real images more difficult to interpret than synthetic ones, it can be intuited that the size of this model is not sufficient for our task.

\subsection{Computer-vision models}

In order to circumvent the lower generalization capabilities of shallow networks, \textcolor{blue}{we consider} models that are successful in computer vision tasks. In this work we focused on the ResNet architecture \cite{he2016deep}. When using models from the literature, there are two possible approaches:
\begin{itemize}[label=\textbullet]
  \item Using the pre-trained weights from another task, to benefit from the rich embedding representations and associated classification layers learned on a much bigger dataset. In this case, the weights can be fine-tuned by using the pre-trained values as initialization. 
  \item Given a sufficient enough database size for the task, it is possible to train from scratch. This is useful in situations where the task is very different than traditional computer visions tasks like the GPR classification problem. 
\end{itemize}
In the following, we denote the first model as \textbf{RFT} \textcolor{blue}{(ResNet Fine-Tuned)} while the second is \textbf{RRT} \textcolor{blue}{(ResNet Re-Trained from scratch)}.

\textcolor{blue}{Such models are appropriate in handling various classification tasks.} However, a problem lies in the very high number of parameters used for the task at hand, making the inference costly compared \textcolor{blue}{in terms of amount of labeled data} to the previous shallow model. To address this issue, one can consider taking advantage of recent approaches based upon second-order statistics, which have shown promising results in computer-vision as well as in other applications \cite{Kobler2022,brooks2019riemannian,shi2023riemannian}.

\subsection{SPD models}

Covariance representations have been shown to be a relevant description when dealing with noisy signals coming from radar systems \cite{OTK12}. Notably, statistical hypothesis testing over second-order modeling have been successful in target detection in GPR \cite{hoarau17}. For classification tasks, building upon \cite{akodad:hal-04263872}, a preliminary study over a binary classification problem has been done in \cite{9884684} using a covariance pooling approach. The idea is to take advantage of the standard convolutional embedding layers that are learned from a computer-vision dataset while employing second-order statistics as a mean to both reduce the dimension of the embedded feature space while also providing additional spatial invariance and better noise handling. 

\textcolor{blue}{As described in \cite{li:is:2017}, we propose to refine the standard convolutional integration layers by backpropagating the loss gradient onto the covariance estimation layers (input tensor construction and estimation) using the matrix backpropagation calculation given in \cite{7410696}. This new strategy adopted in the paper should enable us to better adapt the various layers to GPR images, which are profoundly different from computer vision data.}

\begin{figure*}[t]
	\begin{adjustwidth}{-2cm}{}
		\centering
		\scalebox{0.45}{\input{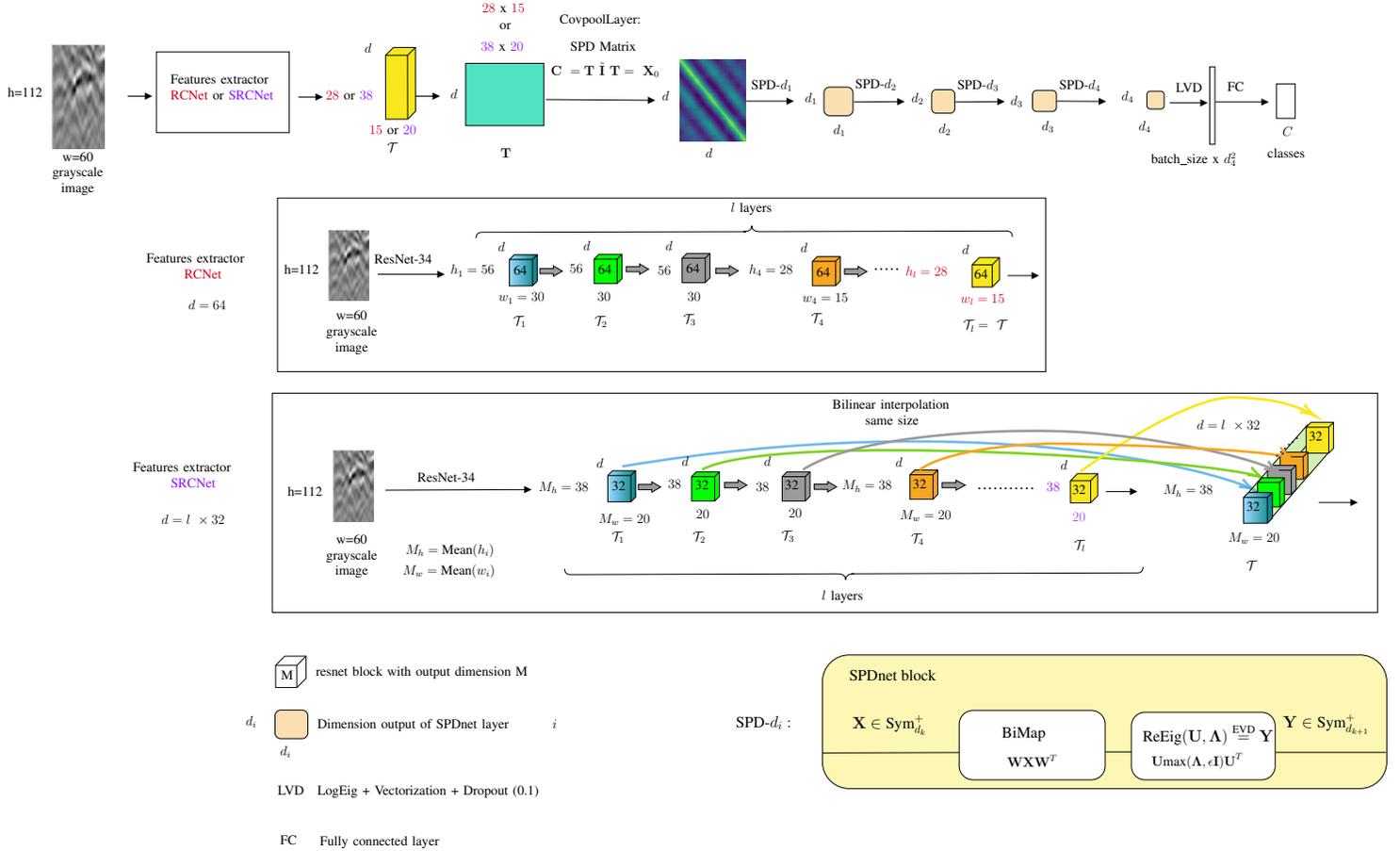}}
        \end{adjustwidth}
	    \caption{\footnotesize Illustration of the two architectures used in this paper. In \textbf{RCNet} (Residual Covariance Network), we take only the last output of the ResNet blocks while in \textbf{SRCNet} (Stacked Residual Covariance Network), we stack the first 32 outputs (to save memory space) of the outputs features by interpolating them to a common size of $38\times 20$. With \textbf{RCNet} (Residual Covariance Network), we have {\color{blue} $h_i=54$ and $w_i=30$ for the first 3 layers and $h_i=28$ and $w_i=15$ for the others.}}
		\label{fig:cdt}
\end{figure*}

\subsection{Proposed models}

Based on those previous works we propose two architectures names \textbf{SRCNet} \textcolor{blue}{(Stacked Residual Covariance Network)} and \textbf{RCNet} \textcolor{blue}{(Residual Covariance Network)} that take advantage of the covariance pooling approach while adding some layers from the recent model \cite{huang2017riemannian} that are specifically designed to handle covariance matrices. The idea behind those layers is to reduce the dimension of the covariance matrices by non-linear dimension reduction while preserving only the information that is relevant for the classification task. The various pipelines used in this paper are illustrated in Figure \ref{fig:cdt}.  Let's describe the different steps of the proposed model:
 
 \textbf{Resnet34 with $l$ layers:} \textcolor{blue}{First, we use the first $l$ layers of a non-pretrained ResNet-34 model\footnote{\textcolor{blue}{Preliminary results have shown that performance is better with an non-pretrained network than with one learned with computer vision data, surely because the first layers generally contain simple features.}}. In this paper, we use $l= 8$, with each layer producing 64 output filters.} We start with a grayscale image $\mathbf{I}\in \mathbb{R}^{h \times w}$ whose initial image size $h=112$ and $w=60$ is taken from \cite{almaimani18}, to provide a basis for comparison in the remainder of this article.
We have considered two scenarios (SRCNet and RCNet), in which we retain a portion of the output from each layer or only the last one.

\textbf{SRCNet:} Starting from our grayscale image, the steps are as follows:
\begin{itemize}
    \item The first $l$ layers of the ResNet34 model each contain $\bar{d}=64$ filters.
    \item  \textcolor{blue}{Only $\bar{d}=32$ filters will be retained for each layer to save memory space.}
    \item Therefore, the total number of filter outputs is given by $d = l \cdot \bar{d}$. 
    \item Let $\mathbf{I}_i \in \mathbb{R}^{h_i \times w_i}$\footnote{Note that each images has a different size for each $i$ which explains the next step to resize all images with an identical size.} with $i \in [1, d]$ represent the set of filter outputs. 
    \item Define $M_h$ as the mean of $h_i$ and $M_w$ as the mean of $w_i$. 
    \item We need to resize the filter outputs $\mathbf{I}_i$ before stacking. For this purpose, we resize $\mathbf{I}_i$ to $\mathbf{\tilde{I}}_{i} \in \mathbb{R}^{M_h \times M_w}$.  
    \item We then stack the resized filter outputs $\mathbf{\tilde{I}}_{i} $ into a tensor $\mathcal{T}$. Thus, $\mathcal{T} = \{ \mathbf{\tilde{I}}_{i} \}_{i \in [1, d]} \in \mathbb{R}^{d \times M_h \times M_w}$.
    
    \item Finally, we reshape $\mathcal{T}$ into $\mathbf{T}$ where $\mathbf{T} \in \mathbb{R}^{d \times M}$ with $M = M_h \times M_w$.
\end{itemize}

 \textbf{RCNet:} \textcolor{blue}{In this alternative approach, the covariance will be calculated without stacking the output filters, but by taking the layers sequentially with all the output filters (and therefore not just 32 in this configuration). In this way, the covariance matrix will be estimated using only the characteristics of the last $l$ layer. In this case, the total number of filters is $d=64$, which leads to $\mathcal{T} = \{ \mathbf{\tilde{I}}_{i} \}_{i \in [1, d]} \in \mathbb{R}^{d \times h_l \times w_l}$. Next, we transform $\mathcal{T}$ into $\mathbf{T}$ where $\mathbf{T} \in \mathbb{R}^{d \times M}$ with $M = h_l \times w_l$. The strategy of this second option aims to explore the potential benefits of residual connections and optimize performance in terms of computation time and memory space.}\\

 \textcolor{blue}{After these steps to build the tensor data $\mathcal{T}$ by either \textbf{SRCNet} or \textbf{RCNet}, we have the following steps:}

\textbf{Covariance Pooling (CovPool) Layer:} \textcolor{blue}{Then a  covariance  pooling layer is added. From $\mathbf{T}$ of \textbf{SRCNet} or \textbf{RCNet} we calculate\footnote{Actually this covariance matrix is the classical Sample Covariance Matrix (SCM) which assumes that the data distribution is Gaussian.}}:
\[\mathbf{C} = \mathbf{T} \bar{\mathbf{I}} \mathbf{T}^\mathrm{T},\]
where $\bar{\mathbf{I}}= \frac{1}{N}\left( \mathbf{I} - \mathbf{1}_{N} \mathbf{1}_{N}^\mathrm{T} \right)$ and $\mathbf{C} \in \ \mathbb{R}^{ d\times d}$ is a SPD Matrix (denoted SPD for Symmetric Positive Definite).\\

\textbf{SPD Net Layer \cite{huang2017riemannian}}:
The obtained covariance matrix $\mathbf{C}$ can then be of very high dimension, and it seems interesting to reduce the data space for better performance. Therefore, we suggest adding convolutional layers adapted to covariance matrices using the framework proposed in \cite{huang2017riemannian}, denoted as SPD Net. 
SPD Net is a model that, for \(k \geq 1\), takes as input a matrix \(\boldsymbol{X}_{k-1} \in \text{Sym}_{d_{k-1}}^{+}\). In our case, we start with \(\mathbf{X}_0 = \mathbf{C} \in \ \mathbb{R}^{ d \times d}\), a symmetric positive definite matrix of size \(d_{k-1}\). The dimensionality of the space is reduced through several BiMap convolution layers and a ReEig regularization layer based on an Eigenvalue Decomposition (EVD). Finally, there is a LogEig layer to perform measurements between covariance matrices in a common space (tangent plane to the Riemannian manifold of SPD matrices taken at the identity). The unknowns in the problem are the convolution matrices \(\boldsymbol{W}_k \in \mathbb{R}_*^{d_k \times d_{k-1}}\), which are low-rank and belong to a Stiefel manifold.

Here are some details about the different layers at each step of the forward propagation:
\begin{itemize}
    \item  BiMap Layer (to generate more compact and discriminative SPD matrices):
    \begin{equation*}
        \mathbf{X}_k = f_b^{(k)}(\mathbf{X}_{k-1};\mathbf{W}_{k})=\mathbf{W}_k\mathbf{X}_{k-1}\mathbf{W}_k^T,
    \end{equation*}
    where $\boldsymbol{X}_{k-1}$ is the input SPD matrix of the \(k\)-th layer, $\boldsymbol{W}_k \in \mathbb{R}_*^{d_k \times d_{k-1}}, (d_k < d_{k-1})$ is the orthonormal transformation matrix (connection weights), $\boldsymbol{X}_k \in \mathbb{R}^{d_k \times d_k}$ is the resulting matrix and  $f_b^{(k)}$ is the function for the $k$-th layer.
    \item  ReEig Layer (to improve discriminative performance, inspired by ReLU):
    \begin{equation*}
        \mathbf{X}_{k} = f_r^{(k)}(\mathbf{X}_{k-1})=\mathbf{U}_{k-1} \max(\epsilon\mathbf{I},\boldsymbol{\Sigma}_{k-1})\mathbf{U}_{k-1}^T,
    \end{equation*}
    where eigenvalue decomposition (EIG) of $\boldsymbol{X}_{k-1}=\mathbf{U}_{k-1} \boldsymbol{\Sigma}_{k-1} \mathbf{U}_{k-1}^T, \epsilon$ is a rectification threshold, $\mathbf{I}$ is an identity matrix and $\max \left(\epsilon \mathbf{I}, \boldsymbol{\Sigma}_{k-1}\right)$ is a diagonal matrix of the corrected eigenvalues in order to stay on the SPD manifold. 
\end{itemize}
\textcolor{blue}{Before the final steps, the BiMap and ReEig layers could be repeated several times in order to find the best representation for the classification.}

\begin{figure}[t]
    \centering
    \begin{tikzpicture}

  \begin{scope}
    \node (input) {$\mathbf{X}$};
    \node[draw, rounded corners=2, right=1cm of input] (f1) {$f_1$};
    \node[below=.3cm of f1.south] (theta1) {$\boldsymbol{\theta}_1$};
    \draw[-stealth, shorten >= 2pt] (input) -- (f1);
    \draw[-stealth, shorten >= 2pt] (theta1) -- (f1);

    \node[right=1cm of f1] (dots) {\dots};
    \draw[-stealth, shorten >= 2pt, shorten <= 2pt] (f1) -- (dots)
    node[midway, above] {$\mathbf{X}_{1}$};

    \node[draw, rounded corners=2, right=1cm of dots] (flminus1) {$f_{\ell-1}$};
    \node[below=.3cm of flminus1.south] (thetalminus1) {$\boldsymbol{\theta}_{\ell-1}$};
    \draw[-stealth, shorten >= 2pt, shorten <= 2pt] (dots) -- (flminus1)
    node[midway, above] {$\mathbf{X}_{\ell-2}$};
    \draw[-stealth, shorten >= 2pt] (thetalminus1) -- (flminus1);

    \node[draw, rounded corners=2, right=1cm of flminus1] (fl) {$f_{\ell}$};
    \node[below=.3cm of fl.south] (thetal) {$\boldsymbol{\theta}_{\ell}$};
    \draw[-stealth, shorten >= 2pt, shorten <= 2pt] (flminus1) -- (fl)
    node[midway, above] {$\mathbf{X}_{\ell-1}$};
    \draw[-stealth, shorten >= 2pt] (thetal) -- (fl);

    \node[right=.3cm of fl] (yhat) {$\hat{\mathbf{y}}$};
    \draw[-stealth, shorten <= 2pt] (fl) -- (yhat);

    \node[draw, rounded corners=2,
    right=.3cm of yhat] (loss) {$\mathcal{L}$};
    \node[above=.3cm of loss.north] (label) {$\mathbf{y}$};
    \draw[-stealth, shorten >= 2pt] (label) -- (loss);
    \draw[-stealth, shorten >= 2pt] (yhat) -- (loss);

  \end{scope}

  \begin{scope}
    \node[draw,
    rounded corners=2,
    below=.75cm of thetal,
    dashed] (gradthetal) {$\nabla f_\ell$};
    \draw[-stealth,
    shorten <= 2pt,
    shorten >= 2pt,
    dashed] (loss.south) -- (gradthetal.east)
    node[pos=.3, below, xshift=.2cm] {$\frac{\partial\mathcal{L}}{\partial\hat{\mathbf{y}}}$};
  \draw[-stealth, shorten >= 2pt, dashed] (thetal) -- (gradthetal);
  \draw[-stealth, shorten >= 2pt, shorten <= 2pt, dashed] (flminus1.east) -- (gradthetal);
  \node[below=.5cm of gradthetal] (newthetal) {$\frac{\partial\mathcal{L}^\ell}{\partial\boldsymbol{\theta}_\ell}$};
  \draw[-stealth, shorten <= 2pt, dashed] (gradthetal) -- (newthetal);

  \node[draw,
  rounded corners=2,
  below=.75cm of thetalminus1,
  dashed] (gradthetalminus1) {$\nabla f_{\ell-1}$};
  \draw[-stealth,
    shorten <= 2pt,
    shorten >= 2pt,
    dashed] (gradthetal) -- (gradthetalminus1)
    node[midway, below, yshift=-4pt] {$\frac{\partial\mathcal{L}^\ell}{\partial \mathbf{X}_{\ell-1}}$};
  \draw[-stealth, shorten >= 2pt, dashed] (thetalminus1) -- (gradthetalminus1);
  \node[below=.5cm of gradthetalminus1] (newthetalminus1) {$\frac{\partial\mathcal{L}^{\ell-1}}{\partial\boldsymbol{\theta}_{\ell-1}}$};
  \draw[-stealth, shorten <= 2pt, dashed] (gradthetalminus1) -- (newthetalminus1);

  \node[left=1cm of gradthetalminus1] (dotsbackprog) {\dots};
  \draw[-stealth,
    shorten <= 2pt,
    shorten >= 2pt,
    dashed] (gradthetalminus1) -- (dotsbackprog)
    node[midway, below, yshift=-4pt] {$\frac{\partial\mathcal{L}^{\ell-1}}{\partial \mathbf{X}_{\ell-2}}$};
  \draw[-stealth, shorten >= 2pt, dashed] (dots.east) -- (gradthetalminus1);

  \node[draw,
  rounded corners=2,
  below=.75cm of theta1,
  dashed] (gradtheta1) {$\nabla f_1$};
  \draw[-stealth, shorten >= 2pt, dashed] (theta1) -- (gradtheta1);
  \node[below=.5cm of gradtheta1] (newthetalminus1) {$\frac{\partial\mathcal{L}^{1}}{\partial\boldsymbol{\theta}_{1}}$};
  \draw[-stealth, shorten <= 2pt, dashed] (gradtheta1) -- (newthetalminus1);
  \draw[-stealth,
    shorten <= 2pt,
    shorten >= 2pt,
    dashed] (dotsbackprog) -- (gradtheta1)
    node[midway, below, yshift=-4pt] {$\frac{\partial\mathcal{L}^{2}}{\partial \mathbf{X}_{1}}$};
  \draw[-stealth, shorten >= 2pt, dashed] (input) -- (gradtheta1);

  \node[
  below=1.8cm of input] (relicat) { };
 \draw[-stealth,
    shorten <= 2pt,
    shorten >= 2pt,
    dashed] (gradtheta1) -- (relicat)
    node[midway, below, yshift=-4pt] {$\frac{\partial\mathcal{L}^{2}}{\partial \mathbf{X}}$};
  \end{scope}

  \draw[thick,
        decorate,
        decoration={brace, raise=1.2cm}] (fl.west) -- (loss.east)
        node[midway, above, yshift=1.3cm] {$\mathcal{L}^\ell$}; 
  \draw[thick,
        decorate,
        decoration={brace, raise=1.75cm}] (flminus1.west) -- (loss.east)
        node[midway, above, yshift=1.85cm] {$\mathcal{L}^{\ell-1}$}; 
  \draw[thick,
        decorate,
        decoration={brace, raise=2.3cm}] (f1.west) -- (loss.east)
        node[midway, above, yshift=2.4cm] {$\mathcal{L}^{1}$}; 

\end{tikzpicture}
    \caption{Matrix backpropagation principle. Solid lines correspond to forward pass and dashed lines to backward pass. $\boldsymbol{\theta}$ are the learned parameters: $\boldsymbol{\theta}_l = \mathbf{W}_l$ for BiMap layers while $\boldsymbol{\theta}=\emptyset$ otherwhise. $\mathbf{X}$ is the tensor after ResNet layers and $\hat{y}$ is the classification output.}
    \label{fig: backprogation principle}
\end{figure}
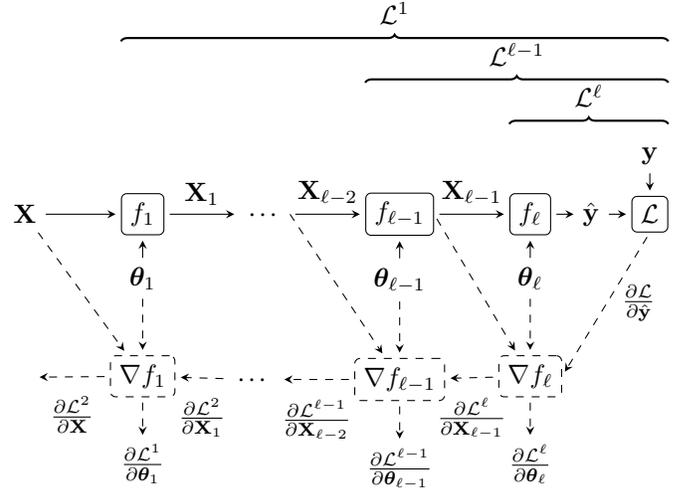

In the final stage of the model, we aim to classify a lower-dimensional discriminant SPD (Symmetric Positive Definite) matrix. To enable the application of traditional fully connected (FC) layers, we first transform the SPD matrix into a feature in Euclidean space. This transformation is performed using the LogEig operator, which calculates the matrix logarithm of the input:
\begin{equation*}
    \mathbf{X}_{k} = f_l^{(k)}(\mathbf{X}_{k-1})=\mathbf{U}_{k-1} \log(\boldsymbol{\Sigma}_{k-1})\mathbf{U}_{k-1}^T,
\end{equation*}
where $\log(\boldsymbol{\Sigma}_{k-1})$ is the diagonal matrix of logarithms of the eigenvalues of the SPD matrix, from the EVD of $\mathbf{X}_{k-1}$ as with ReEig layer. Finally, we vectorize the resulting matrix before the FC layers and we introduce a dropout mechanism to mitigate the risk of overfitting.\\

\textbf{Back-propagation Steps:} Since the main steps are matrix operations, the backward is not classical like in most of deep learning models. In particular, the gradients of the operations based on SVD have been firstly derived in \cite{ionescu2015}. A more stable formula is given in \cite{brooks2019riemannian} and will be used. An illustration of the backpropagation steps can be found in Figure \ref{fig: backprogation principle}. Calculations concerning $\mathbf{W}_k$ differ, as they are based on Riemannian gradient descent on the Stiefel manifold, where $\mathbf{W}_k$ represents an orthonormal matrix. Detailed explanations of this approach can be found in \cite{huang2017riemannian}. Let us summarize hereafter the steps of the backpropagation for the proposed model.

Given loss function $\mathcal{L} : \mathbb{R}^K, \mathbb{R}^K \rightarrow \mathbb{R}$, where $K$ is the number of classes. By denoting $f_l$ the $l$-th layer in the network, we can define the loss starting at this layer as
$$\mathcal{L}^l = \underbrace{\mathcal{L} \circ f_\ell \circ \dots f_{l-1}}_{\mathcal{L}^{l-1}} \circ f_l,$$ 
where $\ell$ is the number of total layers in the network. 

Given those definitions, the backpropagation steps are as follows:
\begin{itemize}

  \item \textit{CovPool Layer}: Let us consider this layer as some $f_l$ in the network, meaning $\mathbf{C}=f_l(\mathbf{T})=\mathbf{T}\bar{\mathbf{I}}\mathbf{T}^T$ and suppose that we have already calculated $\frac{\partial\mathcal{L}^{l+1}}{\partial\mathbf{C}}$. Since there is no parameter to learn, the matrix chain-rule \cite{ionescu2015} on $\mathcal{L}^l(\mathbf{T}) = \mathcal{L}^{l+1}\circ f_l(\mathbf{T})$ yields:
\begin{equation*}
  \langle \frac{\partial \mathcal{L}^{l+1}} {\partial \mathbf{C}}, \mathrm{d}\mathbf{C}\rangle = \langle \frac{\partial\mathcal{L}^l}{\partial \mathbf{T}}, \mathrm{d}\mathbf{T}\rangle,
\end{equation*}
where $\langle \cdot, \cdot \rangle$ is the Frobenius inner product. We have that $\mathrm{d}\mathbf{C}=(\mathrm{d}\mathbf{T})\bar{\mathbf{I}}\mathbf{T}^\mathrm{T} + \mathbf{T}\bar{\mathbf{I}}(\mathrm{d}\mathbf{T})^\mathrm{T}$ and using symmetries on $\bar{\mathbf{I}}$ yields:
\[
  \frac{\partial\mathcal{L}^l}{\partial\mathbf{T}} = 2\bar{\mathbf{I}}\mathbf{T}^\mathrm{T}\frac{\partial\mathcal{L}^{l+1}}{\partial\mathbf{C}}.
\]

  \item \textit{BiMap Layer}: for this layer, we have two gradients to propagate : $\frac{\partial \mathcal{L}^l}{\partial \mathbf{X}_{l-1}}$, the gradient towards the input of the layer and $\frac{\partial\mathcal{L^l}}{\partial\mathbf{W}}$, the gradient to update the weights $\mathbf{W}$ of the bilinear mapping to be learned. \textcolor{blue}{We then use the same matrix back-propagation principles as those presented for the \textit{CovPool} layer, but applied to $f_l:\mathcal{S}^{d_{l-1}} \times \mathcal{O}_{d_{l}\times d_{l-1}}\rightarrow \mathcal{S}^{d_l}$, where $\mathcal{O}_{m\times n}$ is the set of orthonormal matrices of size $m\times n$, and $f_l(\mathbf{X}_{l})=\mathbf{W}\mathbf{X}_{l-1}\mathbf{W}^T$. The gradients are then given by:}
  \begin{equation*}
  \begin{array}{lll}
        \frac{\partial \mathcal{L}^{l}}{\partial \mathbf{X}_{l-1}} &=& \mathbf{W}^\mathrm{T}\frac{\partial \mathcal{L}^{l+1}}{\partial \mathbf{X}_{l}}\mathbf{W} \\
        \frac{\partial \mathcal{L}^{l}}{\partial \mathbf{W}} &=& 2\frac{\partial \mathcal{L}^{l+1}}{\partial \mathbf{X}_{l}}\mathbf{W}\mathbf{X}_{l-1}
  \end{array}.
  \end{equation*}
  One additional thing to take into account is that $\mathbf{W}$ has a special structure. Thus, while doing the gradient step, it is necessary to use a Riemannian retraction operator \cite{boumal2023intromanifolds} to keep the weights on the Stiefel manifold.

\item \textit{ReEig and LogEig Layer}: Since both operations operate on eigenvalue decomposition, we can decompose $f_l$ as $f_l = f_l^{'}\circ\mathrm{eig}$ where $\mathrm{eig}$ means doing the EVD of the input matrix an$ f_l^{'}$ is the operation on the eigenvalues and reconstructing the matrix. The gradient towards input is given by \cite{brooks2019riemannian}:
$$
\begin{aligned}
  \frac{\partial\mathcal{L}^{l}}{\partial\mathbf{X}_{l-1}} =& 2\mathbf{U}_{l-1}\left(\mathbf{P}^{\mathrm{T}}\odot\left(\mathbf{U}_{l-1}^\mathrm{T}\frac{\partial\mathcal{L}^{l+1}\circ f_l^{'}}{\partial\mathbf{U}_{l-1}}\right)_{sym}\right)\mathbf{U}_{l-1}^\mathrm{T} \\
                                                            &+ \mathbf{U}_{l-1}\left(\frac{\partial\mathcal{L}^{l+1}\circ f_l^{'}}{\partial\boldsymbol{\Sigma}_{l-1}}\right)_{diag}\mathbf{U}_{l-1}^\mathrm{T},
\end{aligned}
$$
where $\mathbf{P}$ is a square matrix given by $$\mathbf{P}(i, j) = \begin{cases}\frac{1}{\sigma_i-\sigma_j} & \text{if } i \neq j, \\ 0 & \text{otherwise}\end{cases},$$
and $\sigma_i$ are the eigenvalues of $\mathbf{X}_{l-1}$.

For the ReEig layer, the sub-gradients are given by:
$$
\begin{aligned}
  \frac{\partial\mathcal{L}^{l+1}\circ f_l^{'}}{\partial\mathbf{U}_{l-1}} &= 2\left(\frac{\partial\mathcal{L}^{l+1}}{\partial\mathbf{X}_{l}}\right)_{sym}\mathbf{U}_{l-1}\mathrm{max}(\epsilon\mathbf{I},\boldsymbol{\Sigma}_{l-1}), \\
  \frac{\partial\mathcal{L}^{l+1}\circ f_l^{'}}{\partial\boldsymbol{\Sigma}_{l-1}} &= \mathbf{Q}\mathbf{U}_{l-1}^\mathrm{T}\left(\frac{\partial\mathcal{L}^{l+1}}{\partial\mathbf{X}_{l}}\right)_{sym}\mathbf{U}_{l-1},
\end{aligned}
$$
where $\mathbf{Q}$ is a diagonal matrix with elements: $$\mathbf{Q}(i, i) = \begin{cases}1 & \text{if } \boldsymbol{\Sigma}_{l-1}(i,i) > \epsilon, \\ 0 & \text{otherwise}.\end{cases}$$

For the LogEig layer, the sub-gradients are given by:
$$
\begin{aligned}
  \frac{\partial\mathcal{L}^{l+1}\circ f_l^{'}}{\partial\mathbf{U}_{l-1}} &= 2\left(\frac{\partial\mathcal{L}^{l+1}}{\partial\mathbf{X}_{l}}\right)_{sym}\mathbf{U}_{l-1}\log(\boldsymbol{\Sigma}_{l-1}), \\
  \frac{\partial\mathcal{L}^{l+1}\circ f_l^{'}}{\partial\boldsymbol{\Sigma}_{l-1}} &= \boldsymbol{\Sigma}_{l-1}^{-1}\mathbf{U}_{l-1}^\mathrm{T}\left(\frac{\partial\mathcal{L}^{l+1}}{\partial\mathbf{X}_{l}}\right)_{sym}\mathbf{U}_{l-1}.
\end{aligned}
$$
\end{itemize}

Thanks to all these steps, there is a backpropagation for all the layers from the final fully-connected layers to the ResNet34 convolution layers. This means that contrarily to previous works \cite{li2017second, huang2017riemannian} we make advantage of both the bilinear mapping with learnable weights to obtain a discriminative SPD matrix for classification but also the best convolutions and thus embedding space for the task at hand rather than keeping pre-trained weights on another task.

\textbf{Implementation:} we provide a pyTorch implementation of those steps available at \url{https://github.com/ammarmian/anotherspdnet}. 

\textcolor{blue}{
\textbf{Computation complexity :} Regarding the complexity of the proposed methods (RCNet and SRCNet), the main bottleneck is the computation of the SVD of the covariance input in the BiMap layers which is $O(m^3)$, $m$ being the size of input\footnote{$m=64$ for RCNet and $m=32\times l$ for SCRNet at the first BiMap.}. Intuitively, this makes the RCNet method more computationally attractive than its SRCNet counterpart. With regards to classic methods, it is difficult to conclude on complexity since deriving time complexity is a difficult for deep learning models and highly dependent on the number of parameters as well as the set of hyperparameters used in training. Given the size of model \textbf{S-CNN} though, it is evident that this is the faster method. In order to give an idea of computational cost associated with our models, we consider  hereafter a comparison on time of training and inference of the different models on the same hardware.}

\begin{figure}[H]
    \centering
\begin{tikzpicture}

\definecolor{crimson2143940}{RGB}{214,39,40}
\definecolor{darkorange25512714}{RGB}{255,127,14}
\definecolor{darkslategray38}{RGB}{38,38,38}
\definecolor{forestgreen4416044}{RGB}{44,160,44}
\definecolor{lavender234234242}{RGB}{234,234,242}
\definecolor{lightgray204}{RGB}{204,204,204}
\definecolor{mediumpurple148103189}{RGB}{148,103,189}
\definecolor{orchid227119194}{RGB}{227,119,194}
\definecolor{sienna1408675}{RGB}{140,86,75}
\definecolor{steelblue31119180}{RGB}{31,119,180}

\begin{axis}[
width =\columnwidth,
height=8cm,
tick pos=left,
legend cell align={left},
legend style={
  fill=none,
  draw=none,
  at={(0.03,1.01)},
  anchor=south west,
  font=\footnotesize\selectfont,
  legend columns=3
},
tick align=outside,
x grid style={black!30, dashed},
xlabel=\textcolor{darkslategray38}{Train ratio},
xmajorgrids,
xmin=0.1, xmax=0.9,
xtick style={color=darkslategray38},
y grid style={black!30, dashed},
ylabel=\textcolor{darkslategray38}{Computation time (s)},
ymajorgrids,
ymin=0, ymax=18412.7999999998,
ytick style={color=darkslategray38},
ymode=log
]
\path [draw=white, fill=steelblue31119180, opacity=0.2]
(axis cs:0.1,6)
--(axis cs:0.1,2)
--(axis cs:0.2,2)
--(axis cs:0.3,3)
--(axis cs:0.4,3)
--(axis cs:0.5,3.95)
--(axis cs:0.6,3.95)
--(axis cs:0.7,3.95)
--(axis cs:0.8,3.95)
--(axis cs:0.9,4)
--(axis cs:0.9,7)
--(axis cs:0.9,7)
--(axis cs:0.8,7)
--(axis cs:0.7,7)
--(axis cs:0.6,7)
--(axis cs:0.5,6)
--(axis cs:0.4,6)
--(axis cs:0.3,6)
--(axis cs:0.2,6)
--(axis cs:0.1,6)
--cycle;

\path [draw=white, fill=darkorange25512714, opacity=0.2]
(axis cs:0.1,8)
--(axis cs:0.1,4)
--(axis cs:0.2,6)
--(axis cs:0.3,7)
--(axis cs:0.4,7.95)
--(axis cs:0.5,9)
--(axis cs:0.6,10)
--(axis cs:0.7,12)
--(axis cs:0.8,13)
--(axis cs:0.9,14)
--(axis cs:0.9,20)
--(axis cs:0.9,20)
--(axis cs:0.8,18.3)
--(axis cs:0.7,16)
--(axis cs:0.6,15)
--(axis cs:0.5,13)
--(axis cs:0.4,12)
--(axis cs:0.3,10)
--(axis cs:0.2,9)
--(axis cs:0.1,8)
--cycle;

\path [draw=white, fill=forestgreen4416044, opacity=0.2]
(axis cs:0.1,58.45)
--(axis cs:0.1,27)
--(axis cs:0.2,26)
--(axis cs:0.3,42.95)
--(axis cs:0.4,48.85)
--(axis cs:0.5,43.95)
--(axis cs:0.6,51.85)
--(axis cs:0.7,69.85)
--(axis cs:0.8,64.9)
--(axis cs:0.9,70.95)
--(axis cs:0.9,167)
--(axis cs:0.9,167)
--(axis cs:0.8,134)
--(axis cs:0.7,122)
--(axis cs:0.6,100.15)
--(axis cs:0.5,98.15)
--(axis cs:0.4,125.9)
--(axis cs:0.3,91)
--(axis cs:0.2,66)
--(axis cs:0.1,58.45)
--cycle;

\path [draw=white, fill=crimson2143940, opacity=0.2]
(axis cs:0.1,353.45)
--(axis cs:0.1,173.95)
--(axis cs:0.2,235.75)
--(axis cs:0.3,258.75)
--(axis cs:0.4,385.55)
--(axis cs:0.5,420.95)
--(axis cs:0.6,516.5)
--(axis cs:0.7,535.6)
--(axis cs:0.8,586.15)
--(axis cs:0.9,448.85)
--(axis cs:0.9,1214.3)
--(axis cs:0.9,1214.3)
--(axis cs:0.8,1342.15)
--(axis cs:0.7,1018.3)
--(axis cs:0.6,1028.6)
--(axis cs:0.5,871.4)
--(axis cs:0.4,777.05)
--(axis cs:0.3,523.8)
--(axis cs:0.2,452.2)
--(axis cs:0.1,353.45)
--cycle;

\path [draw=white, fill=mediumpurple148103189, opacity=0.2]
(axis cs:0.1,353.2)
--(axis cs:0.1,159.95)
--(axis cs:0.2,262.65)
--(axis cs:0.3,319.15)
--(axis cs:0.4,386.75)
--(axis cs:0.5,488.4)
--(axis cs:0.6,582.85)
--(axis cs:0.7,532.45)
--(axis cs:0.8,659.95)
--(axis cs:0.9,688.35)
--(axis cs:0.9,1471)
--(axis cs:0.9,1471)
--(axis cs:0.8,1303.65)
--(axis cs:0.7,1160.4)
--(axis cs:0.6,1132.2)
--(axis cs:0.5,953.2)
--(axis cs:0.4,760.45)
--(axis cs:0.3,623.1)
--(axis cs:0.2,490.8)
--(axis cs:0.1,353.2)
--cycle;

\path [draw=white, fill=sienna1408675, opacity=0.2]
(axis cs:0.1,308.4)
--(axis cs:0.1,139.9)
--(axis cs:0.2,157.95)
--(axis cs:0.3,221.6)
--(axis cs:0.4,232.4)
--(axis cs:0.5,436.8)
--(axis cs:0.6,762.55)
--(axis cs:0.7,360.6)
--(axis cs:0.8,321.6)
--(axis cs:0.9,590.8)
--(axis cs:0.9,1667.8)
--(axis cs:0.9,1667.8)
--(axis cs:0.8,502.6)
--(axis cs:0.7,1036.45)
--(axis cs:0.6,1271.6)
--(axis cs:0.5,1105.15)
--(axis cs:0.4,730.3)
--(axis cs:0.3,361.65)
--(axis cs:0.2,252.65)
--(axis cs:0.1,308.4)
--cycle;

\path [draw=white, fill=orchid227119194, opacity=0.2]
(axis cs:0.1,3624)
--(axis cs:0.1,1386.25)
--(axis cs:0.2,2180.25)
--(axis cs:0.3,2228.85)
--(axis cs:0.4,2869.45)
--(axis cs:0.5,3430.95)
--(axis cs:0.6,4478.65)
--(axis cs:0.7,3405.55)
--(axis cs:0.8,3633.65)
--(axis cs:0.9,2504.35)
--(axis cs:0.9,18362.7999999998)
--(axis cs:0.9,18362.7999999998)
--(axis cs:0.8,5446.35)
--(axis cs:0.7,5887.45)
--(axis cs:0.6,13356.4)
--(axis cs:0.5,11963.6)
--(axis cs:0.4,9402.7)
--(axis cs:0.3,5492.95)
--(axis cs:0.2,7854.7)
--(axis cs:0.1,3624)
--cycle;

\addplot [semithick, steelblue31119180, mark=*, mark size=3, mark options={solid}]
table {%
0.1 4.4
0.2 4.67
0.3 5.02
0.4 5.14
0.5 5.28
0.6 5.38
0.7 5.51
0.8 5.83
0.9 6.01
};
\addlegendentry{SVC\_gs\_std\_pca}
\addplot [semithick, darkorange25512714, mark=square*, mark size=3, mark options={solid}]
table {%
0.1 6.91
0.2 8.02
0.3 9.15
0.4 10.41
0.5 11.56
0.6 12.81
0.7 14.13
0.8 16.14
0.9 17.19
};
\addlegendentry{RF\_gs\_std\_pca}
\addplot [semithick, forestgreen4416044, mark=diamond*, mark size=3, mark options={solid}]
table {%
0.1 45.04
0.2 49.45
0.3 70.73
0.4 92.41
0.5 78.23
0.6 82.39
0.7 107.26
0.8 103.88
0.9 122.72
};
\addlegendentry{S-CNN}
\addplot [semithick, crimson2143940, mark=triangle*, mark size=3, mark options={solid}]
table {%
0.1 251.06
0.2 335.51
0.3 367.04
0.4 613.99
0.5 667.64
0.6 820.76
0.7 783.8
0.8 1001.75
0.9 797.47
};
\addlegendentry{RFT}
\addplot [semithick, mediumpurple148103189, mark=triangle*, mark size=3, mark options={solid,rotate=180}]
table {%
0.1 276.41
0.2 382.39
0.3 465.66
0.4 617.72
0.5 731.07
0.6 887.19
0.7 924.11
0.8 1025.28
0.9 1066.67
};
\addlegendentry{RRT}
\addplot [semithick, sienna1408675, mark=pentagon*, mark size=3, mark options={solid}]
table {%
0.1 202.98
0.2 203.98
0.3 291.39
0.4 360.63
0.5 821.87
0.6 1025.73
0.7 547.7
0.8 409.19
0.9 1031.8
};
\addlegendentry{RCNet}
\addplot [semithick, orchid227119194, mark=star, mark size=3, mark options={solid}]
table {%
0.1 2354.07
0.2 3068.04
0.3 3419.12
0.4 4794.7
0.5 7097.84
0.6 9403.32
0.7 5523.54
0.8 5232.22
0.9 10018.05
};
\addlegendentry{SRCNet}
\end{axis}

\end{tikzpicture}
    \caption{\textcolor{blue}{Average runtime w.r.t. ratio of training data set over 100 different seeds. For each method, the line corresponds to the mean of accuracy over all the seeds, and the filled area corresponds to 5-th and 95-th quantiles.}}
    \label{fig:run time}
\end{figure}
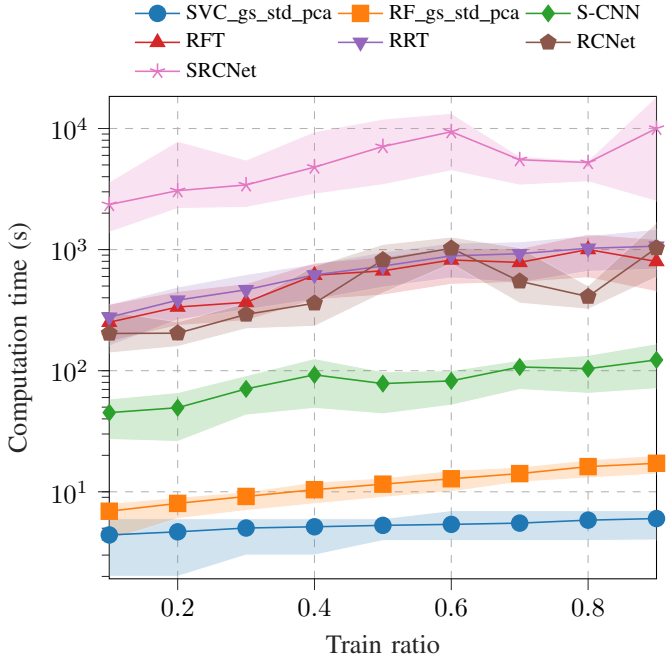

\textcolor{blue}{
In figure \ref{fig:run time}, we show the average run time (training and validation steps) as a function of the ratio of the training dataset for the methods described above and two classical methods (SVM and Random Forests). Full details and the chosen parameters are given in the next section. As might be expected, the classic ML algorithms and the shallow network have the lowest computation times. On the other hand, the SRCNet computation time is clearly the highest. This approach, which consists in stacking all filter outputs, is not at all interesting in terms of computation time. On the other hand, RCNet's computation time is really interesting, since it is most of the time lower than that of ResNet34 -based deep approaches for most training data ratios.}

Let us now present the experimental results obtained with the proposed models in the next sections.

\section{Dataset description}
	\label{sec:dataset}
\begin{figure*}[t]
   \centering
     \scalebox{0.85}{
\begin{tikzpicture}

\definecolor{darkgray176}{RGB}{176,176,176}
\definecolor{green}{RGB}{0,128,0}
\definecolor{lightgray204}{RGB}{204,204,204}
\definecolor{yellow}{RGB}{255,255,0}

\begin{groupplot}[group style={group size=1 by 3,vertical sep=1.5cm},height=6cm,width=20cm]
\nextgroupplot[
tick label style={font=\small},
tick align=outside,
tick pos=left,
x grid style={white!69.0196078431373!black},
xlabel={x (m)},
xmin=-0.5, xmax= 2700,
xtick={0,231,462,693,924,1155,1386,1617,1848,2079,2310,2541,2700},
xticklabels={0.00, 2.33, 4.66, 7.00, 9.33, 11.66, 14.00, 16.33, 18.66, 21.00, 23.33, 24.75, 27.27},
xtick style={color=black},
y dir=reverse,
y grid style={darkgray176},
ylabel={t (ns)},
ymin=-0.5, ymax=900.5,
ytick={0,200,400,600,800},
yticklabels={0.00,21.72,43.46,65.19,86.92},
ytick style={color=black}
]
\addplot graphics [includegraphics cmd=\pgfimage,xmin=-0.5, xmax=
2700, ymin=900.5, ymax=-0.5] {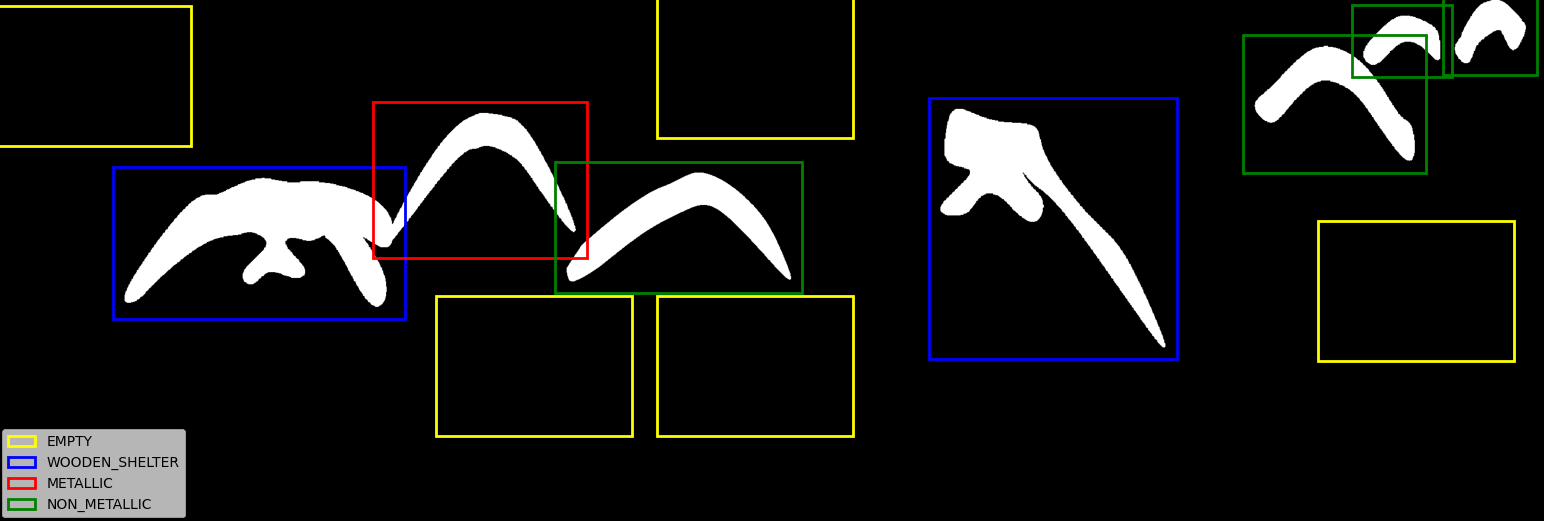};

\nextgroupplot[
tick label style={font=\small},
tick align=outside,
tick pos=left,
x grid style={white!69.0196078431373!black},
xlabel={x (m)},
xmin=-0.5, xmax= 2700,
xtick={0,231,462,693,924,1155,1386,1617,1848,2079,2310,2541,2700},
xticklabels={0.00, 2.33, 4.66, 7.00, 9.33, 11.66, 14.00, 16.33, 18.66, 21.00, 23.33, 24.75, 27.27},
xtick style={color=black},
y dir=reverse,
y grid style={darkgray176},
ylabel={t (ns)},
ymin=-0.5, ymax=900.5,
ytick={0,200,400,600,800},
yticklabels={0.00,21.72,43.46,65.19,86.92},
ytick style={color=black}
]
\addplot graphics [includegraphics cmd=\pgfimage,xmin=-0.5, xmax= 2700, ymin=900.5, ymax=-0.5] {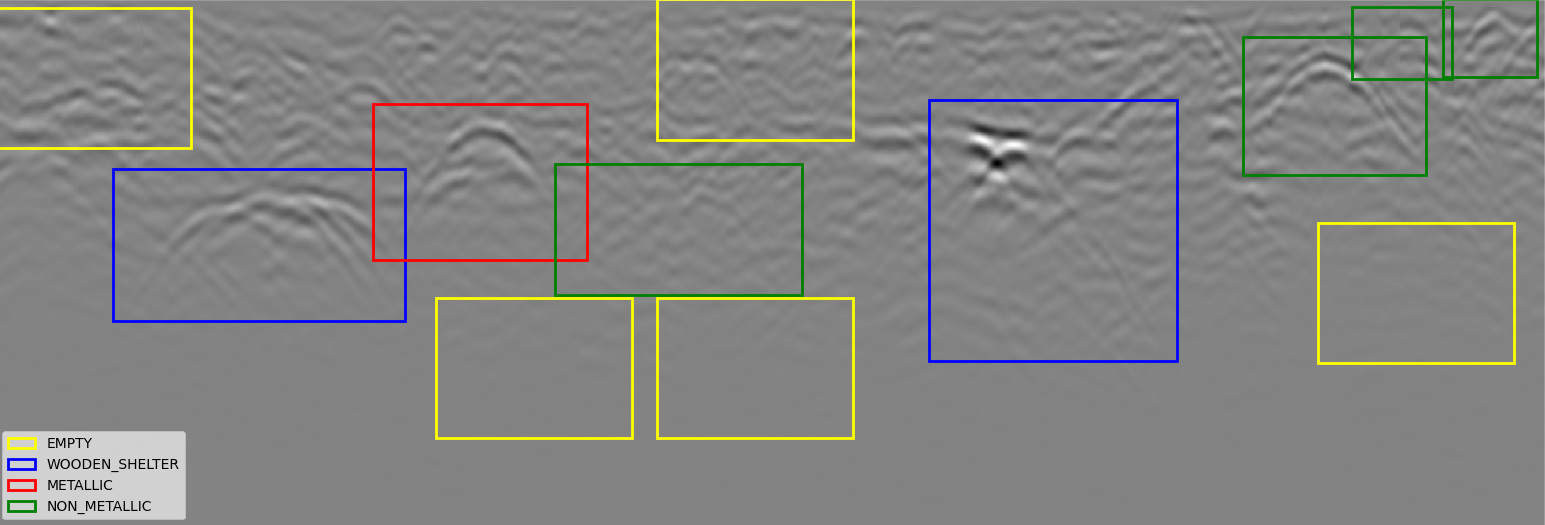};
\end{groupplot}
\end{tikzpicture}}
     \caption{Example for one radargram (bottom) and the corresponding image of the masks. The positions of the rectangles detected on the masks were projected onto the radargrams. The yellow rectangles correspond to a thumbnail of the empty class.}
 \label{fig:rec_rect}  
\end{figure*}

	The database provided by the Geolithe company include 699 medium-sized radargrams of $(R_x, R_y) = (4000, 800)$ pixels, along with all the necessary acquisition information such as radar frequency (200MHz or 350MHz for Geolithe), radar elevation (0cm, 25cm, 50cm, 75cm, 100cm, 150cm), and soil type (wet sand, dry sand, gravel, dry gravel). Each radargram is associated with a mask of the same size. Main preprocessing steps \cite{gallet2022classification} consists to determine the rectangle around each hyperbola in the radargram by using the mask image in order to build a thumbnail image for each target. An example of this preprocessing for one radargram is shown in \ref{fig:rec_rect}. All thumbnails are resized to $(w, h) = (60, 112)$ before to be treated by the different classification methods. We have also created a class denoted "empty" (in yellow in the figure \ref{fig:rec_rect}). \textcolor{blue}{To create this class "empty", we selected 5 rectangles of same size for each radargram of our dataset. The centers of the rectangle are randomly chosen and we check if the rectangle does not overlap with another rectangle containing an object.}

The final database then consists of 1584 thumbnails, classified into four categories: Metallic, Non-Metallic, Wooden Shelters, and Empty. Tables \ref{tab:elv_soil} and \ref{tab:elv_freq} provides more details on the distribution of the four categories according to elevation, soil and frequency. As used classicaly in supervized approaches, we divide our dataset into three distinct parts: a training set consisting of 1108 images, a validation set with 238 images, and finally, a test set also comprising 238 images. The latter will be used to evaluate the accuracy of the different tested models.

\begin{table*}[t]
\centering
\begin{tabular}{r|cccc|cccc|cccc|cccc|c}
     & \multicolumn{4}{c|}{Wooden Shelter} & \multicolumn{4}{c|}{Metallic} & \multicolumn{4}{c|}{Non Metallic} & \multicolumn{4}{c}{Empty} & \multicolumn{1}{c}{\textbf{Total}} \\
    \cline{2-18}
     Soil/Elevation (cm) & \textbf{25} & \textbf{50} & \textbf{75} & \textbf{100} &  \textbf{25} & \textbf{50} & \textbf{75} & \textbf{100}  & \textbf{25} & \textbf{50} & \textbf{75} & \textbf{100} &   \textbf{25} & \textbf{50} & \textbf{75} & \textbf{100} \\
    \hline
    grave & 20 & 16 & 12 & 4 & 13 & 17 & 6& 11 & 16 & 14 & 8 & 4 & 20 & 12 & 18 & 12 & \textbf{203} \\
    \hline
    dry grave & 24 & 18 & 17 & 19 & 18 & 14 & 16 & 6 & 14 & 15 & 14 & 10 & 15 & 19 & 19 & 20 & \textbf{258} \\
    \hline
    sand & 37 & 44 & 48 & 54 & 48 & 40 & 55 & 50 & 40 & 44 & 44 & 48 & 45 & 50 & 45 & 43 & \textbf{735}\\
    \hline
    wet sand  & 18 & 21 & 22 & 22 & 20 & 28 & 22 & 32 & 29 & 26 & 33 & 37 & 19 & 18 & 17 & 24 & \textbf{388} \\
    \cline{1-18}
     Total & 99 & 99 & 99 & 99 & 99 & 99 & 99 & 99 & 99 & 99 & 99 & 99 & 99 & 99 & 99 & 99 & \textbf{1584} \\
\end{tabular}
\caption{Distribution of our database for each soil and depending on the elevation of GPR.}
\label{tab:elv_soil} 
\end{table*}

\begin{table*}[t]
\centering
\begin{tabular}{r|cccc|cccc|cccc|cccc|c}
    & \multicolumn{4}{c|}{Wooden Shelter} & \multicolumn{4}{c|}{Metallic} & \multicolumn{4}{c|}{Non Metallic} & \multicolumn{4}{c|}{Empty} & \multicolumn{1}{c}{\textbf{Total}} \\
    \cline{2-18}
     Frequency/Elevation (cm) & \textbf{25} & \textbf{50} & \textbf{75} & \textbf{100} &  \textbf{25} & \textbf{50} & \textbf{75} & \textbf{100}  & \textbf{25} & \textbf{50} & \textbf{75} & \textbf{100} &   \textbf{25} & \textbf{50} & \textbf{75} & \textbf{100} \\
    \hline
    200 MHZ & 53 & 46 & 54 & 49 & 24 & 13 & 19 & 18 & 26 & 28 & 26 & 20 & 40 & 46 & 45 & 40 & \textbf{547} \\
    \hline
    350 MHZ & 46 & 53 & 45 & 50 & 75 & 86 & 80 & 81 & 73 & 71 & 73 & 79 & 59 & 53 & 54 & 59 & \textbf{1037} \\
    \cline{1-18}
     Total & 99 & 99 & 99 & 99 & 99 & 99 & 99 & 99 & 99 & 99 & 99 & 99 & 99 & 99 & 99 & 99 & \textbf{1584} \\
\end{tabular}
\caption{Distribution of our database for each frequency and depending on the elevation of GPR.}
\label{tab:elv_freq}
\end{table*}

\section{Numerical experiments}
	\label{sec:num_exp}
	\subsection{List of models}

As described in section \ref{sec:som}, we consider in this paper three models with small variation which are recalled in the following:
\begin{itemize}
    \item \textbf{S-CNN}: the shallow architecture developed in \cite{almaimani18} for GPR image classification
    \item \textbf{Resnet34}: a deep network proposed in~\cite{he2016deep} initially applied to computer vision applications. In this section we will consider two models from this architecture:
    \begin{itemize}
        \item \textcolor{blue}{Resnet Re-Trained (\textbf{RRT})}: the model is then trained from scratch, which we initialize the weights randomly.
        \item \textcolor{blue}{Resnet Fine-Tuned (\textbf{RFT})}: the model is fine-tuned which consists in trained by using the pre-trained weights. In this specific model, they are pre-trained from the ImageNet database.
    \end{itemize}
    \item Our proposed models in two configurations:
    \begin{itemize}
        \item \textbf{SRCNet}: all the $l$ first layers (only 32 output filters are selected for each layers to save memory space) are considered to build our tensor. In this case the first size of the SPD matrix is $d_0=256$.
        \item \textbf{RCNet}: only the last layer with these 64 output filters are used to build the tensor. In this case the first size of the SPD matrix is $d_0=64$.
    \end{itemize}
    For these both models, the number of layers of Resnet34 to build the tensor $\mathcal{T}$ is $l=8$. Moreover, SPD Net is designed with 4 consecutive BiMap and ReEig layers where the sizes of each SPD matrices are specified in Table \ref{tab:dimensions spdnet}. The choice of 4 layers is common when using SPD Net \cite{huang2017riemannian} and in particular it is shown in different experiments \cite{dreamnet} that it is useless to take more than 4 consecutive BiMap and ReEig layers. \textcolor{blue}{We also considered models with less than 4 layers as well as other set of output size and settled on this choice by cross-validation with regards to performance.}
\end{itemize}
All models are used with the same parameters each time. The chosen optimizer is SGD with a momentum of 0.9, a batch size of 8, and a learning rate of 0.007.

\begin{table}[h]
    \centering
\begin{tabular}{@{}llllll@{}}
\toprule
       & $d_0$ & $d_2$ & $d_4$ & $d_6$ & $d_8$ \\ \midrule
SRCNet & 256 & 235   & 217   & 179   & 128   \\
RCNet  & 64  & 58    & 54    & 44    & 32    \\ \bottomrule
\end{tabular}
    \caption{Output dimensions of SPDNet layers (BiMap and ReEig).}
    \label{tab:dimensions spdnet}
\end{table}


In the following subsections, we compare all these models with our database described in \ref{sec:dataset}, first by measuring the influence of the number of training data on the classification results, then the robustness in the presence of mislabeled data in the training set, and finally the robustness to data shifts between the training set and the test set. In all experiments, 100 different random generator seeds are used to construct the partition between training and testing data sets. This allows to obtain more representative results irrespective of the initilialization. To showcase results, we decide to show the 5-th, 50-th and 95-th quantiles of the performance metric.

\subsection{Influence of the number of training data}\label{subsec:inf_train}

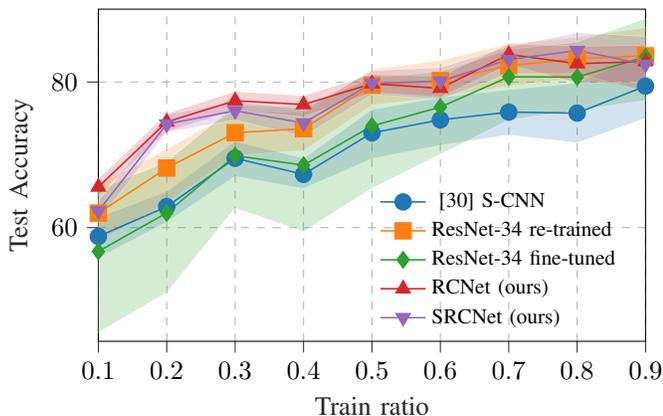
\begin{figure}[h]
    \centering
\begin{tikzpicture}

\definecolor{crimson2143940}{RGB}{214,39,40}
\definecolor{darkorange25512714}{RGB}{255,127,14}
\definecolor{darkslategray38}{RGB}{38,38,38}
\definecolor{forestgreen4416044}{RGB}{44,160,44}
\definecolor{lavender234234242}{RGB}{234,234,242}
\definecolor{lightgray204}{RGB}{204,204,204}
\definecolor{mediumpurple148103189}{RGB}{148,103,189}
\definecolor{steelblue31119180}{RGB}{31,119,180}

\begin{axis}[
width =\columnwidth,
height=6cm,
tick pos=left,
legend cell align={left},
legend style={
  fill=none,
  draw=none,
  at={(0.97,0.01)},
  anchor=south east,
  font=\footnotesize\selectfont
},
tick align=outside,
x grid style={black!30, dashed},
xlabel=\textcolor{darkslategray38}{Train ratio},
xmajorgrids,
xmin=0.1, xmax=0.9,
xtick={0,0.1,0.2,0.3,0.4,0.5,0.6,0.7,0.8,0.9},
xtick style={color=darkslategray38},
y grid style={black!30, dashed},
ylabel=\textcolor{darkslategray38}{Test Accuracy},
ymajorgrids,
ymin=44.388, ymax=90,
ytick style={color=darkslategray38},
]
\path [draw=white, fill=steelblue31119180, opacity=0.2]
(axis cs:0.1,61.437)
--(axis cs:0.1,55.813)
--(axis cs:0.2,60.714)
--(axis cs:0.3,67.03)
--(axis cs:0.4,65.3295)
--(axis cs:0.5,69.44)
--(axis cs:0.6,71.2745)
--(axis cs:0.7,72.69)
--(axis cs:0.8,71.6055)
--(axis cs:0.9,75)
--(axis cs:0.9,83.8125)
--(axis cs:0.9,83.8125)
--(axis cs:0.8,79.87)
--(axis cs:0.7,78.99)
--(axis cs:0.6,77.92)
--(axis cs:0.5,76.26)
--(axis cs:0.4,69.561)
--(axis cs:0.3,71.899)
--(axis cs:0.2,65.004)
--(axis cs:0.1,61.437)
--cycle;

\path [draw=white, fill=darkorange25512714, opacity=0.2]
(axis cs:0.1,65.92)
--(axis cs:0.1,58.053)
--(axis cs:0.2,63.394)
--(axis cs:0.3,69.883)
--(axis cs:0.4,70.5585)
--(axis cs:0.5,76.77)
--(axis cs:0.6,77.904)
--(axis cs:0.7,79.41)
--(axis cs:0.8,80.5)
--(axis cs:0.9,78.75)
--(axis cs:0.9,87.5)
--(axis cs:0.9,87.5)
--(axis cs:0.8,86.16)
--(axis cs:0.7,85.311)
--(axis cs:0.6,82.9855)
--(axis cs:0.5,81.82)
--(axis cs:0.4,75.84)
--(axis cs:0.3,76.049)
--(axis cs:0.2,70.82)
--(axis cs:0.1,65.92)
--cycle;

\path [draw=white, fill=forestgreen4416044, opacity=0.2]
(axis cs:0.1,65.255)
--(axis cs:0.1,45.538)
--(axis cs:0.2,51.037)
--(axis cs:0.3,62.637)
--(axis cs:0.4,59.4395)
--(axis cs:0.5,65.3495)
--(axis cs:0.6,70.03)
--(axis cs:0.7,74.727)
--(axis cs:0.8,76.1)
--(axis cs:0.9,77.4375)
--(axis cs:0.9,88.75)
--(axis cs:0.9,88.75)
--(axis cs:0.8,85.53)
--(axis cs:0.7,84.471)
--(axis cs:0.6,81.437)
--(axis cs:0.5,80.6105)
--(axis cs:0.4,76.68)
--(axis cs:0.3,76.778)
--(axis cs:0.2,69.0975)
--(axis cs:0.1,65.255)
--cycle;

\path [draw=white, fill=crimson2143940, opacity=0.2]
(axis cs:0.1,66.76)
--(axis cs:0.1,64.373)
--(axis cs:0.2,73.174)
--(axis cs:0.3,76.031)
--(axis cs:0.4,75.21)
--(axis cs:0.5,78.527)
--(axis cs:0.6,77.904)
--(axis cs:0.7,82.35)
--(axis cs:0.8,79.839)
--(axis cs:0.9,80)
--(axis cs:0.9,85.0625)
--(axis cs:0.9,85.0625)
--(axis cs:0.8,84.91)
--(axis cs:0.7,85.29)
--(axis cs:0.6,80.76)
--(axis cs:0.5,81.0725)
--(axis cs:0.4,78.15)
--(axis cs:0.3,78.74)
--(axis cs:0.2,75.71)
--(axis cs:0.1,66.76)
--cycle;

\path [draw=white, fill=mediumpurple148103189, opacity=0.2]
(axis cs:0.1,63.677)
--(axis cs:0.1,60.863)
--(axis cs:0.2,73.014)
--(axis cs:0.3,74.77)
--(axis cs:0.4,72.48)
--(axis cs:0.5,78.255)
--(axis cs:0.6,77.6)
--(axis cs:0.7,81.09)
--(axis cs:0.8,81.13)
--(axis cs:0.9,78.6875)
--(axis cs:0.9,86.25)
--(axis cs:0.9,86.25)
--(axis cs:0.8,86.8215)
--(axis cs:0.7,84.891)
--(axis cs:0.6,82.02)
--(axis cs:0.5,81.5825)
--(axis cs:0.4,76.2705)
--(axis cs:0.3,77.309)
--(axis cs:0.2,75.574)
--(axis cs:0.1,63.677)
--cycle;

\addplot [semithick, steelblue31119180, mark=*, mark size=3, mark options={solid}]
table {%
0.1 58.7831
0.2 62.8812
0.3 69.5266
0.4 67.3161
0.5 73.0066
0.6 74.8013
0.7 75.861
0.8 75.7294
0.9 79.4375
};
\addlegendentry{\cite{almaimani18} S-CNN}
\addplot [semithick, darkorange25512714, mark=square*, mark size=3, mark options={solid}]
table {%
0.1 61.9877
0.2 68.1817
0.3 73.0271
0.4 73.5888
0.5 79.5588
0.6 80.1898
0.7 82.224
0.8 83.2017
0.9 83.5875
};
\addlegendentry{ResNet-34 re-trained}
\addplot [semithick, forestgreen4416044, mark=diamond*, mark size=3, mark options={solid}]
table {%
0.1 56.7362
0.2 62.022
0.3 69.8792
0.4 68.5811
0.5 73.9601
0.6 76.552
0.7 80.7078
0.8 80.6733
0.9 83.4625
};
\addlegendentry{ResNet-34 fine-tuned}
\addplot [semithick, crimson2143940, mark=triangle*, mark size=3, mark options={solid}]
table {%
0.1 65.5798
0.2 74.5222
0.3 77.4314
0.4 76.9068
0.5 79.7613
0.6 79.1387
0.7 83.7906
0.8 82.5161
0.9 82.8125
};
\addlegendentry{RCNet (ours)}
\addplot [semithick, mediumpurple148103189, mark=triangle*, mark size=3, mark options={solid,rotate=180}]
table {%
0.1 62.2788
0.2 74.2267
0.3 76.0494
0.4 74.3658
0.5 79.8134
0.6 80.124
0.7 83.0808
0.8 84.3251
0.9 82.4145
};
\addlegendentry{SRCNet (ours)}
\end{axis}

\end{tikzpicture}
    \caption{Results of test accuracy w.r.t to training dataset percentage over 100 different seeds. For each method, the line corresponds to the mean of accuracy over all the seeds, and the filled area corresponds to 5-th and 95-th quantiles.}
    \label{fig:cdt1}
\end{figure}

\textcolor{blue}{First, we study the influence of the number of training data since we want to design classification models which can provide good performance with a limited training dataset.} The results of test accuracy w.r.t training ratio are shown in Figure \ref{fig:cdt1}. We can observe that our proposed approaches, \textbf{SRCNet} and \textbf{RCNet} both outperform the classical approaches at any given training ratio and have lower variability over seeds. The \textbf{Resnet34} approaches perform better than \cite{almaimani18} especially when the number of training data increases. The retraining approach, \textbf{RRT}, gives better results compared to the fine-tuning approach, \textbf{RFT}, in particular when the training ratio becomes smaller.

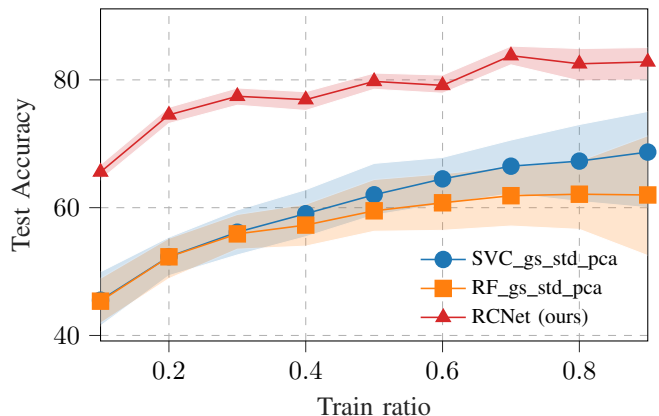
\begin{figure}[h]
    \centering
\begin{tikzpicture}

\definecolor{crimson2143940}{RGB}{214,39,40}
\definecolor{darkorange25512714}{RGB}{255,127,14}
\definecolor{darkslategray38}{RGB}{38,38,38}
\definecolor{forestgreen4416044}{RGB}{44,160,44}
\definecolor{lavender234234242}{RGB}{234,234,242}
\definecolor{lightgray204}{RGB}{204,204,204}
\definecolor{mediumpurple148103189}{RGB}{148,103,189}
\definecolor{steelblue31119180}{RGB}{31,119,180}

\begin{axis}[
width =\columnwidth,
height=6cm,
tick pos=left,
legend cell align={left},
legend style={
  fill=none,
  draw=none,
  at={(0.97,0.01)},
  anchor=south east,
  font=\footnotesize\selectfont
},
tick align=outside,
x grid style={black!30, dashed},
xlabel=\textcolor{darkslategray38}{Train ratio},
xmajorgrids,
xmin=0.1, xmax=0.9,
xtick style={color=darkslategray38},
y grid style={black!30, dashed},
ylabel=\textcolor{darkslategray38}{Test Accuracy},
ymajorgrids,
ymin=39.1382363253857, ymax=91.1124649368864,
ytick style={color=darkslategray38}
]
\path [draw=white, fill=steelblue31119180, opacity=0.2]
(axis cs:0.1,49.9789621318373)
--(axis cs:0.1,41.5007012622721)
--(axis cs:0.2,49.3611987381703)
--(axis cs:0.3,52.5945945945946)
--(axis cs:0.4,55.4516806722689)
--(axis cs:0.5,58.8383838383838)
--(axis cs:0.6,60.5678233438486)
--(axis cs:0.7,62.1848739495798)
--(axis cs:0.8,60.9748427672956)
--(axis cs:0.9,60)
--(axis cs:0.9,75.0625)
--(axis cs:0.9,75.0625)
--(axis cs:0.8,73.0188679245283)
--(axis cs:0.7,70.5882352941177)
--(axis cs:0.6,67.8706624605678)
--(axis cs:0.5,66.9318181818182)
--(axis cs:0.4,62.8256302521008)
--(axis cs:0.3,59.6396396396396)
--(axis cs:0.2,55.3785488958991)
--(axis cs:0.1,49.9789621318373)
--cycle;

\path [draw=white, fill=darkorange25512714, opacity=0.2]
(axis cs:0.1,48.9551192145863)
--(axis cs:0.1,41.9284712482468)
--(axis cs:0.2,48.8880126182965)
--(axis cs:0.3,53.5045045045045)
--(axis cs:0.4,53.9810924369748)
--(axis cs:0.5,56.3005050505051)
--(axis cs:0.6,56.4511041009464)
--(axis cs:0.7,57.1218487394958)
--(axis cs:0.8,56.6037735849057)
--(axis cs:0.9,52.5)
--(axis cs:0.9,71.25)
--(axis cs:0.9,71.25)
--(axis cs:0.8,67.2955974842767)
--(axis cs:0.7,66.4285714285714)
--(axis cs:0.6,65.2996845425868)
--(axis cs:0.5,64.4065656565656)
--(axis cs:0.4,60.5357142857143)
--(axis cs:0.3,58.9369369369369)
--(axis cs:0.2,55.205047318612)
--(axis cs:0.1,48.9551192145863)
--cycle;

\path [draw=white, fill=crimson2143940, opacity=0.2]
(axis cs:0.1,66.76)
--(axis cs:0.1,64.373)
--(axis cs:0.2,73.174)
--(axis cs:0.3,76.031)
--(axis cs:0.4,75.21)
--(axis cs:0.5,78.527)
--(axis cs:0.6,77.904)
--(axis cs:0.7,82.35)
--(axis cs:0.8,79.839)
--(axis cs:0.9,80)
--(axis cs:0.9,85.0625)
--(axis cs:0.9,85.0625)
--(axis cs:0.8,84.91)
--(axis cs:0.7,85.29)
--(axis cs:0.6,80.76)
--(axis cs:0.5,81.0725)
--(axis cs:0.4,78.15)
--(axis cs:0.3,78.74)
--(axis cs:0.2,75.71)
--(axis cs:0.1,66.76)
--cycle;

\addplot [semithick, steelblue31119180, mark=*, mark size=3, mark options={solid}]
table {%
0.1 45.5217391304348
0.2 52.3375394321767
0.3 56.163963963964
0.4 59.0588235294118
0.5 62.020202020202
0.6 64.5078864353312
0.7 66.4915966386555
0.8 67.2893081761006
0.9 68.7
};
\addlegendentry{SVC\_gs\_std\_pca}
\addplot [semithick, darkorange25512714, mark=square*, mark size=3, mark options={solid}]
table {%
0.1 45.367461430575
0.2 52.3091482649842
0.3 55.8972972972973
0.4 57.2731092436975
0.5 59.5075757575758
0.6 60.7634069400631
0.7 61.8739495798319
0.8 62.1069182389937
0.9 61.9875
};
\addlegendentry{RF\_gs\_std\_pca}
\addplot [semithick, crimson2143940, mark=triangle*, mark size=3, mark options={solid}]
table {%
0.1 65.5798
0.2 74.5222
0.3 77.4314
0.4 76.9068
0.5 79.7613
0.6 79.1387
0.7 83.7906
0.8 82.5161
0.9 82.8125
};
\addlegendentry{RCNet (ours)}
\end{axis}

\end{tikzpicture}
    \caption{\textcolor{blue}{Comparison of results compared to classical machine learning models. Test accuracy w.r.t to training dataset percentage over 100 different seeds. For each method, the line corresponds to the mean of accuracy over all the seeds, and the filled area corresponds to 5-th and 95-th quantiles.}}
    \label{fig:classif ml classic}
\end{figure}

\textcolor{blue}{In order to showcase the benefits or using CNNs compared to more classical machine learning methods, we also compare these results to Support Vector Machine (SVM) and Random Forests which are reported to be used in GPR applications\cite{zhou18, 9200094}. 
To that end, we used standard image processing pipeline
  : we vectorize the images and standardize to have zero mean and variance $1$. Then Principal Component Analysis (PCA) is used to reduce dimensionality with a threshold of 95\% of explained variance. Concerning the methods, we use:}
{\color{blue}
\begin{itemize}
    \item[$\bullet$] SVM classifier using a radial basis function (RBF) kernel whose hyperparameters are optimized through grid search.
    \item[$\bullet$] RF classifer : We used Gini index \cite{breiman2001random} for node splitting. Again the hyperparameters are tuned using a grid search approach.
\end{itemize}

The results are presented in Figure \ref{fig:classif ml classic} where they are compared to our best performing model RCNet. This comparison allows to show that deep learning models are better performing in our scenario.
}

\subsection{Robustness to mislabeled Data}\label{subsec:rob_miss}

\begin{figure}[h]
    \centering
\begin{tikzpicture}

\definecolor{crimson2143940}{RGB}{214,39,40}
\definecolor{darkorange25512714}{RGB}{255,127,14}
\definecolor{darkslategray38}{RGB}{38,38,38}
\definecolor{forestgreen4416044}{RGB}{44,160,44}
\definecolor{lavender234234242}{RGB}{234,234,242}
\definecolor{lightgray204}{RGB}{204,204,204}
\definecolor{mediumpurple148103189}{RGB}{148,103,189}
\definecolor{steelblue31119180}{RGB}{31,119,180}

\begin{axis}[
width =\columnwidth,
height=5cm,
tick pos=left,
legend cell align={left},
legend style={
  fill=none,
  draw=none,
  at={(0,-0.02)},
  anchor=south west,
  font=\footnotesize\selectfont
},
tick align=outside,
x grid style={black!30, dashed},
xlabel=\textcolor{darkslategray38}{Mislabelling percentage},
xmajorgrids,
xmin=0, xmax=0.2,
xtick={0,0.05,0.1,0.15,0.2,0.25,0.3,0.35,0.4},
xtick style={color=darkslategray38},
y grid style={black!30, dashed},
ylabel=\textcolor{darkslategray38}{Test Accuracy},
ymajorgrids,
ymin=22, ymax=89.344,
ytick style={color=darkslategray38},
]
\path [draw=white, fill=steelblue31119180, opacity=0.2]
(axis cs:0,78.99)
--(axis cs:0,72.69)
--(axis cs:0.05,62.5885)
--(axis cs:0.1,24.79)
--(axis cs:0.15,24.79)
--(axis cs:0.2,24.349)
--(axis cs:0.25,24.79)
--(axis cs:0.3,24.79)
--(axis cs:0.35,24.769)
--(axis cs:0.4,25.21)
--(axis cs:0.4,59.681)
--(axis cs:0.4,59.681)
--(axis cs:0.35,59.681)
--(axis cs:0.3,63.471)
--(axis cs:0.25,63.051)
--(axis cs:0.2,66.432)
--(axis cs:0.15,66.39)
--(axis cs:0.1,69.75)
--(axis cs:0.05,73.971)
--(axis cs:0,78.99)
--cycle;

\path [draw=white, fill=darkorange25512714, opacity=0.2]
(axis cs:0,85.311)
--(axis cs:0,79.41)
--(axis cs:0.05,75.609)
--(axis cs:0.1,71.85)
--(axis cs:0.15,67.209)
--(axis cs:0.2,65.088)
--(axis cs:0.25,59.24)
--(axis cs:0.3,58.358)
--(axis cs:0.35,53.738)
--(axis cs:0.4,48.74)
--(axis cs:0.4,60.122)
--(axis cs:0.4,60.122)
--(axis cs:0.35,63.891)
--(axis cs:0.3,67.23)
--(axis cs:0.25,71.451)
--(axis cs:0.2,75.63)
--(axis cs:0.15,76.47)
--(axis cs:0.1,79.431)
--(axis cs:0.05,81.51)
--(axis cs:0,85.311)
--cycle;

\path [draw=white, fill=forestgreen4416044, opacity=0.2]
(axis cs:0,84.471)
--(axis cs:0,74.727)
--(axis cs:0.05,69.309)
--(axis cs:0.1,64.269)
--(axis cs:0.15,58.358)
--(axis cs:0.2,58.4)
--(axis cs:0.25,56.279)
--(axis cs:0.3,51.68)
--(axis cs:0.35,45.779)
--(axis cs:0.4,41.936)
--(axis cs:0.4,61.8025)
--(axis cs:0.4,61.8025)
--(axis cs:0.35,64.752)
--(axis cs:0.3,68.07)
--(axis cs:0.25,70.611)
--(axis cs:0.2,73.551)
--(axis cs:0.15,75.651)
--(axis cs:0.1,78.171)
--(axis cs:0.05,81.111)
--(axis cs:0,84.471)
--cycle;

\path [draw=white, fill=crimson2143940, opacity=0.2]
(axis cs:0,85.29)
--(axis cs:0,82.35)
--(axis cs:0.05,81.51)
--(axis cs:0.1,80.25)
--(axis cs:0.15,79.41)
--(axis cs:0.2,78.99)
--(axis cs:0.25,78.15)
--(axis cs:0.3,77.31)
--(axis cs:0.35,76.029)
--(axis cs:0.4,73.95)
--(axis cs:0.4,78.171)
--(axis cs:0.4,78.171)
--(axis cs:0.35,79.851)
--(axis cs:0.3,81.51)
--(axis cs:0.25,81.531)
--(axis cs:0.2,82.77)
--(axis cs:0.15,83.19)
--(axis cs:0.1,84.03)
--(axis cs:0.05,85.29)
--(axis cs:0,85.29)
--cycle;

\path [draw=white, fill=mediumpurple148103189, opacity=0.2]
(axis cs:0,84.891)
--(axis cs:0,81.09)
--(axis cs:0.05,79.809)
--(axis cs:0.1,77.31)
--(axis cs:0.15,73.53)
--(axis cs:0.2,73.95)
--(axis cs:0.25,71.43)
--(axis cs:0.3,69.75)
--(axis cs:0.35,65.529)
--(axis cs:0.4,65.13)
--(axis cs:0.4,71.451)
--(axis cs:0.4,71.451)
--(axis cs:0.35,72.27)
--(axis cs:0.3,76.05)
--(axis cs:0.25,77.331)
--(axis cs:0.2,80.25)
--(axis cs:0.15,79.431)
--(axis cs:0.1,82.371)
--(axis cs:0.05,83.631)
--(axis cs:0,84.891)
--cycle;

\addplot [semithick, steelblue31119180, mark=*, mark size=3, mark options={solid}]
table {%
0 75.861
0.05 68.2164
0.1 54.1439
0.15 43.0423
0.2 37.8234
0.25 37.6888
0.3 38.7267
0.35 35.9367
0.4 38.9949
};

\addplot [semithick, darkorange25512714, mark=square*, mark size=3, mark options={solid}]
table {%
0 82.224
0.05 78.8766
0.1 75.7854
0.15 72.1859
0.2 70.548
0.25 65.128
0.3 62.1303
0.35 58.2417
0.4 54.3723
};
\addplot [semithick, forestgreen4416044, mark=diamond*, mark size=3, mark options={solid}]
table {%
0 80.7078
0.05 76.5372
0.1 71.308
0.15 66.5977
0.2 65.409
0.25 63.7401
0.3 60.8194
0.35 56.7257
0.4 53.8139
};
\addplot [semithick, crimson2143940, mark=triangle*, mark size=3, mark options={solid}]
table {%
0 83.7906
0.05 83.505
0.1 82.077
0.15 81.258
0.2 80.9136
0.25 79.7754
0.3 79.578
0.35 77.9022
0.4 76.2348
};
\addplot [semithick, mediumpurple148103189, mark=triangle*, mark size=3, mark options={solid,rotate=180}]
table {%
0 83.0808
0.05 81.7704
0.1 79.9392
0.15 77.1042
0.2 77.6082
0.25 74.727
0.3 72.5294
0.35 69.5064
0.4 68.1698
};
\end{axis}

\end{tikzpicture}
    \caption{Results of test accuracy w.r.t training dataset mislabelling percentage. For each method, the line corresponds to the mean of accuracy over all the seeds, and the filled area corresponds to 5-th and 95-th quantiles. The legends are the same as in Fig. \ref{fig:cdt1}.}
    \label{fig:cdt2}
\end{figure}
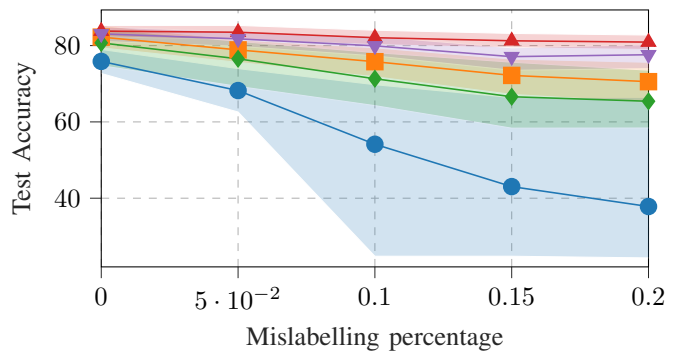

The second simulation studies the effect to have mislabeled data in the training data set. Actually it can be difficult to correctly labeled the hyperbola in particular because of the low Signal to Noise Ratio (SNR) in radargram. For this simulation, we introduce a variable level of mislabeled data, 0 to 20\% of error, in the training set. 
The results of test accuracy w.r.t mislabelling percentage are presented in Figure \ref{fig:cdt2}. We have comparable dynamic than the previous experiment with better performance and robustness of \textbf{RCNet} and \textbf{SRCNet} (with a better result when considering only the last layer for the tensor construction). The case of \cite{almaimani18} is interesting as it shows that shallow models performance decreases quickly as soon as there are a few mislabelled data. \textcolor{blue}{Like in the previous simulation}, we observe that our approaches have the least variability over the seeds. In conclusion, this result is in line with the analysis made in \cite{collas2022robustgeometricmetriclearning}, which also shows that covariance matrix utilization brings great robustness to mislabeled data in metric learning methods.

\subsection{Robustness to data shift}\label{subsec:rob_mod}

\textcolor{blue}{In the case of GPR data, data are collected in different configurations: different frequencies and elevations for RADAR, different soils and weather conditions. As it is difficult (if not impossible) to have labeled data for all these configurations, it is very important to build models with the property of being robust to possible changes between training and test data. These transformations are also known in many applications and are referred to as data shifts. To study the behavior of the models proposed in this article, we consider 4 scenarios:}
\begin{itemize}[label=*]
    \item Scenario A: the training and validation sets include images obtained using a RADAR located at an altitude of 75 cm or 100 cm, while the test set is made up of images acquired at an altitude of 50 cm.
    \item Scenario B: the frequency is chosen at 200 MHz for the training and validation sets, while the test set uses only data obtained with a frequency of 350 MHz.
    \item Scenario C: the training and validation sets are made up of data acquired in dry gravel, whereas the test set uses gravel.
    \item Scenario D: same scenario as C, with wet sand for the training and validation sets, and dry sand for the test set.
\end{itemize}
The distribution and number of elements in the sets for each scenario are detailed in table \ref{tab:da}.

\begin{table}[h]
    \centering
    \begin{tabular}{@{}llllll@{}}
    \toprule
       & Train & Val & Test \\ \midrule
        A: ELV 75,100 vs 50 & 634 & 79   & 79     \\
        B: FRQ 350 vs 200 & 394 & 49   & 49   \\
        C: GRD Dry Gravel  vs Gravel & 394 & 49   & 49    \\
        D: GRD Wet Sand vs Dry Sand  & 672  & 84   & 84    \\ 
    \bottomrule
    \end{tabular}
    \caption{Distribution of the training, validation and test sets for the 4 scenario.}
    \vspace{-1.5em}
    \label{tab:da}
\end{table}


\subsubsection{Scenario A}\label{subsecsec:elv}

\begin{figure}[h]
    \centering
\begin{tikzpicture}

\definecolor{crimson}{RGB}{220,20,60}
\definecolor{darkorange}{RGB}{255,140,0}
\definecolor{darkslategray38}{RGB}{38,38,38}
\definecolor{forestgreen}{RGB}{34,139,34}
\definecolor{lavender234234242}{RGB}{234,234,242}
\definecolor{mediumpurple}{RGB}{147,112,219}
\definecolor{steelblue}{RGB}{70,130,180}

\begin{axis}[
width = \columnwidth,
height=5cm,
tick pos=left,
tick align=outside,
x grid style={white},
xmajorgrids,
xmin=-0.5, xmax=4.5,
xtick style={color=darkslategray38},
xtick={0,1,2,3,4},
xticklabels={
  S-CNN,
  RRT,
  RFT,
  RCNet,
  SRCNet
},
y grid style={white},
ylabel=\textcolor{darkslategray38}{Test Accuracy (\%)},
ymin=39.6815, ymax=85.6285,
ytick style={color=darkslategray38}
]
\path [draw=black, fill=steelblue]
(axis cs:-0.25,62.03)
--(axis cs:0.25,62.03)
--(axis cs:0.25,65.82)
--(axis cs:-0.25,65.82)
--(axis cs:-0.25,62.03)
--cycle;
\addplot [black]
table {%
0 62.03
0 58.23
};
\addplot [black]
table {%
0 65.82
0 70.89
};
\addplot [black]
table {%
-0.125 58.23
0.125 58.23
};
\addplot [black]
table {%
-0.125 70.89
0.125 70.89
};
\addplot [black, mark=o, mark size=3, mark options={solid,fill opacity=0}, only marks]
table {%
0 54.43
0 72.15
0 73.42
};
\path [draw=black, fill=darkorange]
(axis cs:0.75,70.89)
--(axis cs:1.25,70.89)
--(axis cs:1.25,75.95)
--(axis cs:0.75,75.95)
--(axis cs:0.75,70.89)
--cycle;
\addplot [black]
table {%
1 70.89
1 67.09
};
\addplot [black]
table {%
1 75.95
1 82.28
};
\addplot [black]
table {%
0.875 67.09
1.125 67.09
};
\addplot [black]
table {%
0.875 82.28
1.125 82.28
};
\path [draw=black, fill=forestgreen]
(axis cs:1.75,62.03)
--(axis cs:2.25,62.03)
--(axis cs:2.25,73.42)
--(axis cs:1.75,73.42)
--(axis cs:1.75,62.03)
--cycle;
\addplot [black]
table {%
2 62.03
2 49.37
};
\addplot [black]
table {%
2 73.42
2 78.48
};
\addplot [black]
table {%
1.875 49.37
2.125 49.37
};
\addplot [black]
table {%
1.875 78.48
2.125 78.48
};
\addplot [black, mark=o, mark size=3, mark options={solid,fill opacity=0}, only marks]
table {%
2 41.77
};
\path [draw=black, fill=crimson]
(axis cs:2.75,77.22)
--(axis cs:3.25,77.22)
--(axis cs:3.25,79.75)
--(axis cs:2.75,79.75)
--(axis cs:2.75,77.22)
--cycle;
\addplot [black]
table {%
3 77.22
3 74.68
};
\addplot [black]
table {%
3 79.75
3 83.54
};
\addplot [black]
table {%
2.875 74.68
3.125 74.68
};
\addplot [black]
table {%
2.875 83.54
3.125 83.54
};
\addplot [black, mark=o, mark size=3, mark options={solid,fill opacity=0}, only marks]
table {%
3 73.42
};
\path [draw=black, fill=mediumpurple]
(axis cs:3.75,72.15)
--(axis cs:4.25,72.15)
--(axis cs:4.25,74.68)
--(axis cs:3.75,74.68)
--(axis cs:3.75,72.15)
--cycle;
\addplot [black]
table {%
4 72.15
4 69.62
};
\addplot [black]
table {%
4 74.68
4 77.22
};
\addplot [black]
table {%
3.875 69.62
4.125 69.62
};
\addplot [black]
table {%
3.875 77.22
4.125 77.22
};
\addplot [black, mark=o, mark size=3, mark options={solid,fill opacity=0}, only marks]
table {%
4 68.35
4 68.35
4 68.35
4 68.35
};
\addplot [black]
table {%
-0.25 63.29
0.25 63.29
};
\addplot [black]
table {%
0.75 74.68
1.25 74.68
};
\addplot [black]
table {%
1.75 67.72
2.25 67.72
};
\addplot [black]
table {%
2.75 78.48
3.25 78.48
};
\addplot [black]
table {%
3.75 73.42
4.25 73.42
};
\end{axis}

\end{tikzpicture}
    \caption{Results of test accuracy results for scenario A \textcolor{blue}{(ELV 75,100 vs 50).}}
    \label{fig:da_elv}
\end{figure}
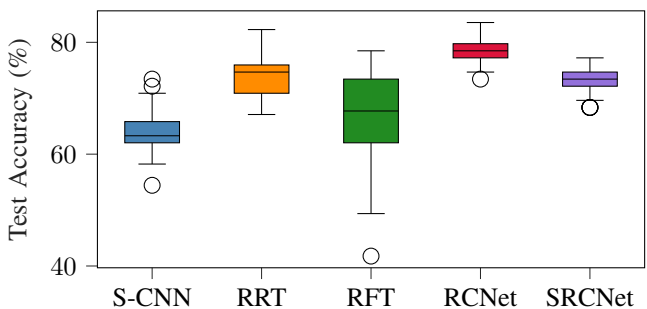

In Figure \ref{fig:da_elv} we represent five boxplots given the accuracy performances of the different models. We easily concluded that our models, in particular \textbf{RCNet} performance is almost unchanged compared to the classical case, are particular robust to this transformation. In this case, main transformations are scaling but the shape of the hyperbola slightly changes. As expected, the shallow model is the less robust. We also noticed that the variability over the seeds is very small with the models built from covariance matrix. This result is also in line with those obtained in  \cite{Collas2023} where covariance matrix are used to classify crops in Satellite Image Times Series. 

\subsubsection{Scenario B}\label{subsecsec:frq}

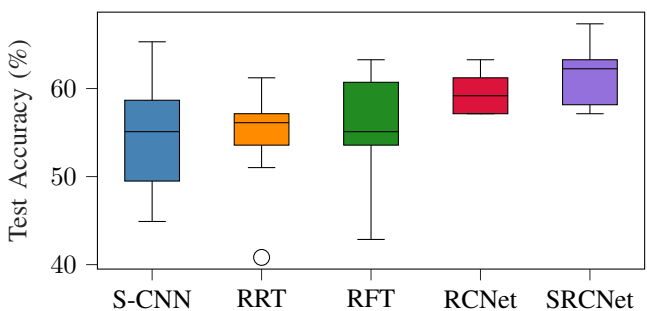
\begin{figure}[h]
    \centering
\begin{tikzpicture}

\definecolor{crimson}{RGB}{220,20,60}
\definecolor{darkorange}{RGB}{255,140,0}
\definecolor{darkslategray38}{RGB}{38,38,38}
\definecolor{forestgreen}{RGB}{34,139,34}
\definecolor{lavender234234242}{RGB}{234,234,242}
\definecolor{mediumpurple}{RGB}{147,112,219}
\definecolor{steelblue}{RGB}{70,130,180}

\begin{axis}[
width = \columnwidth,
height=5cm,
tick pos=left,
tick align=outside,
x grid style={white},
xmajorgrids,
xmin=-0.5, xmax=4.5,
xtick style={color=darkslategray38},
xtick={0,1,2,3,4},
xticklabels={
  S-CNN,
  RRT,
  RFT,
  RCNet,
  SRCNet
},
y grid style={white},
ylabel=\textcolor{darkslategray38}{Test Accuracy (\%)},
ymin=39.4935, ymax=68.6765,
ytick style={color=darkslategray38}
]
\path [draw=black, fill=steelblue]
(axis cs:-0.25,49.49)
--(axis cs:0.25,49.49)
--(axis cs:0.25,58.67)
--(axis cs:-0.25,58.67)
--(axis cs:-0.25,49.49)
--cycle;
\addplot [black]
table {%
0 49.49
0 44.9
};
\addplot [black]
table {%
0 58.67
0 65.31
};
\addplot [black]
table {%
-0.125 44.9
0.125 44.9
};
\addplot [black]
table {%
-0.125 65.31
0.125 65.31
};
\path [draw=black, fill=darkorange]
(axis cs:0.75,53.57)
--(axis cs:1.25,53.57)
--(axis cs:1.25,57.14)
--(axis cs:0.75,57.14)
--(axis cs:0.75,53.57)
--cycle;
\addplot [black]
table {%
1 53.57
1 51.02
};
\addplot [black]
table {%
1 57.14
1 61.22
};
\addplot [black]
table {%
0.875 51.02
1.125 51.02
};
\addplot [black]
table {%
0.875 61.22
1.125 61.22
};
\addplot [black, mark=o, mark size=3, mark options={solid,fill opacity=0}, only marks]
table {%
1 40.82
};
\path [draw=black, fill=forestgreen]
(axis cs:1.75,53.57)
--(axis cs:2.25,53.57)
--(axis cs:2.25,60.71)
--(axis cs:1.75,60.71)
--(axis cs:1.75,53.57)
--cycle;
\addplot [black]
table {%
2 53.57
2 42.86
};
\addplot [black]
table {%
2 60.71
2 63.27
};
\addplot [black]
table {%
1.875 42.86
2.125 42.86
};
\addplot [black]
table {%
1.875 63.27
2.125 63.27
};
\path [draw=black, fill=crimson]
(axis cs:2.75,57.14)
--(axis cs:3.25,57.14)
--(axis cs:3.25,61.22)
--(axis cs:2.75,61.22)
--(axis cs:2.75,57.14)
--cycle;
\addplot [black]
table {%
3 57.14
3 57.14
};
\addplot [black]
table {%
3 61.22
3 63.27
};
\addplot [black]
table {%
2.875 57.14
3.125 57.14
};
\addplot [black]
table {%
2.875 63.27
3.125 63.27
};
\path [draw=black, fill=mediumpurple]
(axis cs:3.75,58.16)
--(axis cs:4.25,58.16)
--(axis cs:4.25,63.27)
--(axis cs:3.75,63.27)
--(axis cs:3.75,58.16)
--cycle;
\addplot [black]
table {%
4 58.16
4 57.14
};
\addplot [black]
table {%
4 63.27
4 67.35
};
\addplot [black]
table {%
3.875 57.14
4.125 57.14
};
\addplot [black]
table {%
3.875 67.35
4.125 67.35
};
\addplot [black]
table {%
-0.25 55.1
0.25 55.1
};
\addplot [black]
table {%
0.75 56.12
1.25 56.12
};
\addplot [black]
table {%
1.75 55.1
2.25 55.1
};
\addplot [black]
table {%
2.75 59.18
3.25 59.18
};
\addplot [black]
table {%
3.75 62.245
4.25 62.245
};
\end{axis}

\end{tikzpicture}
    \caption{Results of test accuracy results for scenario B \textcolor{blue}{(FRQ 350 vs 200 394 49 49).}}
    \label{fig:da_frq}
\end{figure}

The box diagrams are now shown in figure \ref{fig:da_frq}. As in the previous scenario, the models we propose outperform \textcolor{blue}{the classical models}. But in this case, the best algorithm is \textbf{SRCNet} and we notice a sharp degradation in the performance of all approaches. The number of transformations between the two sets is then too high. This is because the frequency is linked to the penetration of the wave into the ground, as well as to the possible resolution which results in very different radargrams. 

\subsubsection{Scenario C and D}\label{subsecsec:grv}

\begin{figure}[h]
    \centering
\begin{tikzpicture}

\definecolor{crimson}{RGB}{220,20,60}
\definecolor{darkorange}{RGB}{255,140,0}
\definecolor{darkslategray38}{RGB}{38,38,38}
\definecolor{forestgreen}{RGB}{34,139,34}
\definecolor{lavender234234242}{RGB}{234,234,242}
\definecolor{mediumpurple}{RGB}{147,112,219}
\definecolor{steelblue}{RGB}{70,130,180}

\begin{axis}[
width = \columnwidth,
height=5cm,
tick pos=left,
tick align=outside,
x grid style={white},
xmajorgrids,
xmin=-0.5, xmax=4.5,
xtick style={color=darkslategray38},
xtick={0,1,2,3,4},
xticklabels={
  S-CNN,
  RRT,
  RFT,
  RCNet,
  SRCNet
},
y grid style={white},
ylabel=\textcolor{darkslategray38}{Test Accuracy (\%)},
ymin=36.331, ymax=90.209,
ytick style={color=darkslategray38}
]
\path [draw=black, fill=steelblue]
(axis cs:-0.25,65.31)
--(axis cs:0.25,65.31)
--(axis cs:0.25,71.43)
--(axis cs:-0.25,71.43)
--(axis cs:-0.25,65.31)
--cycle;
\addplot [black]
table {%
0 65.31
0 57.14
};
\addplot [black]
table {%
0 71.43
0 77.55
};
\addplot [black]
table {%
-0.125 57.14
0.125 57.14
};
\addplot [black]
table {%
-0.125 77.55
0.125 77.55
};
\addplot [black, mark=o, mark size=3, mark options={solid,fill opacity=0}, only marks]
table {%
0 55.1
};
\path [draw=black, fill=darkorange]
(axis cs:0.75,63.27)
--(axis cs:1.25,63.27)
--(axis cs:1.25,69.9)
--(axis cs:0.75,69.9)
--(axis cs:0.75,63.27)
--cycle;
\addplot [black]
table {%
1 63.27
1 57.14
};
\addplot [black]
table {%
1 69.9
1 77.55
};
\addplot [black]
table {%
0.875 57.14
1.125 57.14
};
\addplot [black]
table {%
0.875 77.55
1.125 77.55
};
\addplot [black, mark=o, mark size=3, mark options={solid,fill opacity=0}, only marks]
table {%
1 81.63
};
\path [draw=black, fill=forestgreen]
(axis cs:1.75,57.14)
--(axis cs:2.25,57.14)
--(axis cs:2.25,69.39)
--(axis cs:1.75,69.39)
--(axis cs:1.75,57.14)
--cycle;
\addplot [black]
table {%
2 57.14
2 38.78
};
\addplot [black]
table {%
2 69.39
2 77.55
};
\addplot [black]
table {%
1.875 38.78
2.125 38.78
};
\addplot [black]
table {%
1.875 77.55
2.125 77.55
};
\path [draw=black, fill=crimson]
(axis cs:2.75,79.59)
--(axis cs:3.25,79.59)
--(axis cs:3.25,83.67)
--(axis cs:2.75,83.67)
--(axis cs:2.75,79.59)
--cycle;
\addplot [black]
table {%
3 79.59
3 75.51
};
\addplot [black]
table {%
3 83.67
3 87.76
};
\addplot [black]
table {%
2.875 75.51
3.125 75.51
};
\addplot [black]
table {%
2.875 87.76
3.125 87.76
};
\path [draw=black, fill=mediumpurple]
(axis cs:3.75,71.43)
--(axis cs:4.25,71.43)
--(axis cs:4.25,75.51)
--(axis cs:3.75,75.51)
--(axis cs:3.75,71.43)
--cycle;
\addplot [black]
table {%
4 71.43
4 65.31
};
\addplot [black]
table {%
4 75.51
4 81.63
};
\addplot [black]
table {%
3.875 65.31
4.125 65.31
};
\addplot [black]
table {%
3.875 81.63
4.125 81.63
};
\addplot [black]
table {%
-0.25 67.35
0.25 67.35
};
\addplot [black]
table {%
0.75 67.35
1.25 67.35
};
\addplot [black]
table {%
1.75 64.29
2.25 64.29
};
\addplot [black]
table {%
2.75 81.63
3.25 81.63
};
\addplot [black]
table {%
3.75 73.47
4.25 73.47
};
\end{axis}

\end{tikzpicture}
    \caption{Results of test accuracy results for scenario C \textcolor{blue}{(GRD Dry Gravel vs Gravel).}}
    \label{fig:da_grave}
\end{figure}
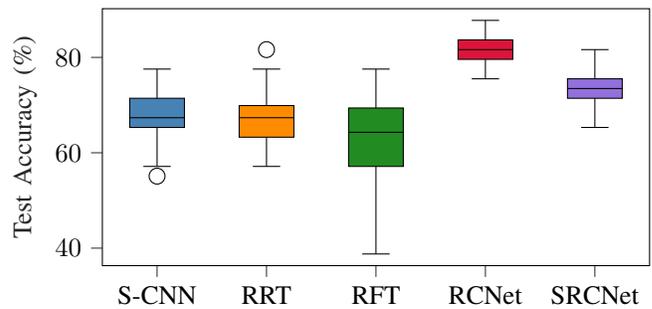

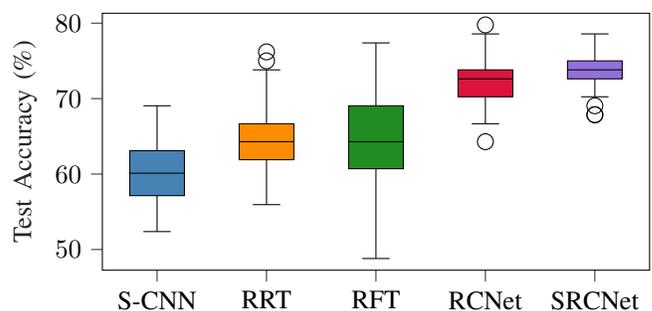
\begin{figure}[h]
    \centering
\begin{tikzpicture}

\definecolor{crimson}{RGB}{220,20,60}
\definecolor{darkorange}{RGB}{255,140,0}
\definecolor{darkslategray38}{RGB}{38,38,38}
\definecolor{forestgreen}{RGB}{34,139,34}
\definecolor{lavender234234242}{RGB}{234,234,242}
\definecolor{mediumpurple}{RGB}{147,112,219}
\definecolor{steelblue}{RGB}{70,130,180}

\begin{axis}[
width = \columnwidth,
height=5cm,
tick pos=left,
tick align=outside,
x grid style={white},
xmajorgrids,
xmin=-0.5, xmax=4.5,
xtick style={color=darkslategray38},
xtick={0,1,2,3,4},
xticklabels={
  S-CNN,
  RRT,
  RFT,
  RCNet,
  SRCNet
},
y grid style={white},
ylabel=\textcolor{darkslategray38}{Test Accuracy (\%)},
ymin=47.2625, ymax=81.3075,
ytick style={color=darkslategray38}
]
\path [draw=black, fill=steelblue]
(axis cs:-0.25,57.14)
--(axis cs:0.25,57.14)
--(axis cs:0.25,63.1)
--(axis cs:-0.25,63.1)
--(axis cs:-0.25,57.14)
--cycle;
\addplot [black]
table {%
0 57.14
0 52.38
};
\addplot [black]
table {%
0 63.1
0 69.05
};
\addplot [black]
table {%
-0.125 52.38
0.125 52.38
};
\addplot [black]
table {%
-0.125 69.05
0.125 69.05
};
\path [draw=black, fill=darkorange]
(axis cs:0.75,61.9)
--(axis cs:1.25,61.9)
--(axis cs:1.25,66.67)
--(axis cs:0.75,66.67)
--(axis cs:0.75,61.9)
--cycle;
\addplot [black]
table {%
1 61.9
1 55.95
};
\addplot [black]
table {%
1 66.67
1 73.81
};
\addplot [black]
table {%
0.875 55.95
1.125 55.95
};
\addplot [black]
table {%
0.875 73.81
1.125 73.81
};
\addplot [black, mark=o, mark size=3, mark options={solid,fill opacity=0}, only marks]
table {%
1 76.19
1 75
};
\path [draw=black, fill=forestgreen]
(axis cs:1.75,60.71)
--(axis cs:2.25,60.71)
--(axis cs:2.25,69.05)
--(axis cs:1.75,69.05)
--(axis cs:1.75,60.71)
--cycle;
\addplot [black]
table {%
2 60.71
2 48.81
};
\addplot [black]
table {%
2 69.05
2 77.38
};
\addplot [black]
table {%
1.875 48.81
2.125 48.81
};
\addplot [black]
table {%
1.875 77.38
2.125 77.38
};
\path [draw=black, fill=crimson]
(axis cs:2.75,70.24)
--(axis cs:3.25,70.24)
--(axis cs:3.25,73.81)
--(axis cs:2.75,73.81)
--(axis cs:2.75,70.24)
--cycle;
\addplot [black]
table {%
3 70.24
3 66.67
};
\addplot [black]
table {%
3 73.81
3 78.57
};
\addplot [black]
table {%
2.875 66.67
3.125 66.67
};
\addplot [black]
table {%
2.875 78.57
3.125 78.57
};
\addplot [black, mark=o, mark size=3, mark options={solid,fill opacity=0}, only marks]
table {%
3 64.29
3 79.76
};
\path [draw=black, fill=mediumpurple]
(axis cs:3.75,72.62)
--(axis cs:4.25,72.62)
--(axis cs:4.25,75)
--(axis cs:3.75,75)
--(axis cs:3.75,72.62)
--cycle;
\addplot [black]
table {%
4 72.62
4 70.24
};
\addplot [black]
table {%
4 75
4 78.57
};
\addplot [black]
table {%
3.875 70.24
4.125 70.24
};
\addplot [black]
table {%
3.875 78.57
4.125 78.57
};
\addplot [black, mark=o, mark size=3, mark options={solid,fill opacity=0}, only marks]
table {%
4 67.86
4 69.05
4 67.86
};
\addplot [black]
table {%
-0.25 60.115
0.25 60.115
};
\addplot [black]
table {%
0.75 64.29
1.25 64.29
};
\addplot [black]
table {%
1.75 64.29
2.25 64.29
};
\addplot [black]
table {%
2.75 72.62
3.25 72.62
};
\addplot [black]
table {%
3.75 73.81
4.25 73.81
};
\end{axis}

\end{tikzpicture}
    \caption{Results of test accuracy results for scenario D \textcolor{blue}{(GRD Wet Sand vs Dry Sand).}}
    \label{fig:da_sand}
\end{figure}

The boxplots for the scenario C are shown in \ref{fig:da_grave} while those for the scenario D are given in \ref{fig:da_sand}. In both cases, the second order
deep learning models give better results and lower variabilities. We can conclude that if we want to obtain robust performances, it is clear that our approaches are better suited than shallow models or classical deep learning models as \textbf{Resnet34}.

\section{Conclusion}

In this paper, we have proposed a new deep learning model based on second-order moments to classify buried objects from the hyperbola thumbnails obtained with a classical GPR system. The proposed model is the concatenation of several models: the first is composed of the first layers of a classical CNN and is used to obtain a covariance matrix from the outputs of convolutional filters, while the second is composed of specific layers to classify SPD matrices. These models are tested on a database composed of several radargrams and compared with shallow models and conventional CNNs typically used in computer vision applications. Our approach gives better results, particularly when the number of training data decreases and in the presence of mislabeled data. We also illustrated the value of second-order deep learning models when training data and test sets are obtained from different weather modes or considerations.

\textcolor{blue}{Thanks to this last analysis, we believe that our algorithms will perform well for other GPR datasets. In addition, the proposed architecture will be interesting to test on other RADAR data classification applications (e.g. remote sensing), but also in medical applications (e.g. EEG). As in GPR, these fields of application use very noisy data and do not have a large number of labeled data.}

\textcolor{blue}{The main extension of this work will be to develop a robust, high-performance model that combines the detection, localization and classification stages. In particular, we want to integrate second-order layers into the classical fast-RCNN network \cite{girshick2014rcnn,ren2015fasterrcnn,girshick2015fastrcnn}. Indeed, the first blocks of this network is very close to our models, since it constructs a data tensor from convolutional layers. As in the proposed work in the current paper, these new layers could make it possible to obtain a more robust and efficient network with a low number of labeled data.}

\bibliographystyle{IEEEbib}
\bibliography{biblio}

\end{document}